\documentclass[10pt]{article}
\usepackage[letterpaper,margin=0.72in]{geometry}
\usepackage{amsmath,amssymb}
\usepackage{graphicx}
\usepackage{booktabs}
\usepackage{longtable}
\usepackage{array}
\usepackage{calc}
\usepackage{enumitem}
\usepackage{xurl}
\usepackage{natbib}
\usepackage[hidelinks]{hyperref}
\usepackage{caption}
\usepackage{parskip}
\setlist{nosep,leftmargin=*}
\hypersetup{pdftitle={Linglong Agent Service: A Runtime for Centralized Agent Reasoning and Distributed Execution},pdfauthor={ZHANG Hua and Min Jiang}}
\providecommand{\tightlist}{}

\providecommand{\real}[1]{#1}
\begin{document}
\begin{center}
{\Large\bfseries Linglong Agent Service: A Runtime for Centralized Agent Reasoning and Distributed Execution\par}
\vspace{0.5em}
{\normalsize ZHANG Hua\textsuperscript{1}\quad Min Jiang\textsuperscript{2}\par}
\vspace{0.25em}
{\small \textsuperscript{1}Independent Researcher\\
\textsuperscript{2}Jiangnan University\\
\href{mailto:mz24cn@hotmail.com}{mz24cn@hotmail.com}\\
\href{mailto:minjiang@jiangnan.edu.cn}{minjiang@jiangnan.edu.cn}\par}
\vspace{0.35em}
{\small English arXiv draft V2B --- November 2026\par}
\end{center}
\begin{abstract}
LLM agents can use remote tools and cloud sandboxes, but integrating reasoning, session state, and heterogeneous execution sites remains a runtime design problem. We present Linglong Agent Service (agentsvc), a self-hosted HTTP service and embeddable Python runtime that treats execution site as a configurable dimension alongside model, tools, and context. A parent service retains inference orchestration, tool descriptions, and conversation history; each inference request selects one child environment for file, shell, terminal, and MCP operations. Direct HTTP and child-initiated WebSocket tunnels carry the same application interfaces, including portable request context and centrally managed updates. The normal core path requires no third-party Python packages, with an optional Windows pseudo-terminal dependency. We further present online\_debugger as the process-scale special case of execution-site separation: an instrumented program submits exception evidence, executes session-scoped inspection tools inside the incident process, and resumes the same diagnostic session. Its default prompt requires failing-then-passing reproduction evidence for source changes, while runtime patch activation and failed-call recovery remain unimplemented. We explain the design through constrained expected-utility reasoning, compare concrete mechanisms with Hermes Agent, Pi, Codex, and Claude Code, and report implementation and regression-test evidence. The current evidence establishes executable contracts, not performance advantages, repair accuracy, or production recovery. The source code is open source under the MIT license at \url{https://github.com/mz24cn/agents}.
\end{abstract}
\textbf{Keywords:} LLM agent; agent runtime; remote execution; execution site; client-side tools; exception diagnosis; self-healing
\hypertarget{introduction}{%
\section{Introduction}\label{introduction}}

An LLM agent acts through tools, and every tool executes somewhere:
inside a process, in a filesystem namespace, under a permission
boundary, and on a machine with particular resources. Existing agents
already reach beyond their own host through SSH, remote MCP servers, and
hosted sandboxes. The remaining engineering question is how to preserve
a coherent agent service when its actions must occur across
infrastructure with different network, installation, and state
constraints.

Consider a coding agent that operates a GPU server behind NAT. The model
provider and conversation history can remain centrally managed, while
commands, files, persistent terminals, and tool-specific resources
belong to the GPU server. Moving only the shell leaves other operations
and their histories to be integrated separately. Running a separate full
agent on every machine is also possible, but can duplicate configuration
and fragment the operator's view. Neither outcome is inevitable; they
motivate an explicit contract between centralized orchestration and
distributed execution.

Linglong Agent Service, abbreviated \textbf{agentsvc}, explores that
contract in a self-hosted runtime \citep{agentsvc}. The implementation
is open source under the MIT license and publicly available at
\url{https://github.com/mz24cn/agents}. Its design vision is to
concentrate agent capability in one place and project execution across
IT infrastructure. Here, \emph{projection} means selecting where
compatible tools execute while retaining centrally managed reasoning and
session state. Infrastructure-wide reach is a design goal subject to
Python availability, network reachability, authorization,
operating-system support, and tool compatibility. It is not a claim that
every device has been supported or tested. Centralized reasoning refers
to inference orchestration: model weights may reside at a local or
remote provider.

The organizing model is

\[
\text{Agent} = \text{Model} \times \text{Tools} \times
\text{Context} \times \text{Execution Site}.
\]

The product notation expresses configurable composition, not
mathematical independence. Execution site is explicit because changing
it should not require changing model-facing tool names, rebuilding an
agent definition, or relocating conversation history. Compatibility
conditions remain: the selected model must support the interaction, the
child must implement the required tools, and paths must resolve in the
intended namespace.

The same argument extends below machine granularity. When an ordinary
program catches an exception, evidence in its traceback and frames may
be useful to an external coding agent. A shell on the same host does not
automatically provide access to those process objects. The included
\textbf{online\_debugger} captures evidence and exposes inspection
operations as client-side tools executed by the incident process. This
makes it a process-scale special case of execution-site separation. It
shares the architectural principle with infrastructure remote execution,
while using a different transport and continuation contract.

This paper makes four contributions:

\begin{enumerate}
\def\labelenumi{\arabic{enumi}.}
\tightlist
\item
  A compositional agent runtime and a design rationale that connects
  centralized session assets, low-dependency deployment, and runtime
  configuration under an explicit expected-utility criterion.
\item
  A remote-execution design with request-scoped site selection, portable
  execution context, parent-side orchestration, and common application
  interfaces over direct HTTP and reverse WebSocket transports.
\item
  A client-side tool contract and its realization in exception-triggered
  diagnosis and source repair, with an explicit separation between
  reproduction evidence and runtime recovery.
\item
  A mechanism-level measurement program for that design, built on a
  reusable validity contract: seven experiment families on routing and
  transport cost, completion decisions in persistent terminals,
  interactive layer traversal, tool-semantics parity across routing
  arms, injected failure and authorization behavior, and execution-site
  switching, plus a bounded comparison with four agent systems and an
  evaluation agenda that separates mechanism correctness from task
  effectiveness.
\end{enumerate}

The implementation baseline is commit
\texttt{07d3d23a24f356bf2e59527b762ab66611c34f27}, dated November 8,
2026 \citep{agentsvc}. Reported regression results come from the
verification record for that baseline. The measurements reported here
are mechanism-level. They were obtained with scripted drivers and a
deterministic model endpoint on a local lab harness, a deployed
instance, and a three-host interactive chain, at pilot repetition and
without a model making task decisions. We still have no controlled
model-repair study, no token-benefit measurement, and no evidence of
general production recovery to report.

\hypertarget{design-rationale}{%
\section{Design Rationale}\label{design-rationale}}

\hypertarget{constraints-and-near-term-assumptions}{%
\subsection{Constraints and near-term
assumptions}\label{constraints-and-near-term-assumptions}}

We begin with six constraints and working assumptions. First, common
model APIs permit application orchestration to be separated from model
execution. A model request can be approximated as a stochastic mapping
from messages and tool descriptions to a message and tool calls;
provider-side caches or response state do not remove the application's
responsibility for sessions and execution. Second, the agent's direct
effects in the systems studied here pass through tools with concrete
execution sites. Third, context windows, output limits, latency, cost,
and permissions jointly constrain tasks. Fourth, installation, network,
and privilege requirements affect whether a runtime can enter an
existing environment. Fifth, some diagnostic evidence is transient and
can be lost when a process exits or shared state changes. Sixth, some
choices require a human or another external authority.

The corresponding near-term assumptions are continued diversity of model
providers, persistent heterogeneity in infrastructure, and continuing
demand for long-running agent tasks. These are revisable judgments, not
measured laws. They favor replaceable protocol adapters, explicit
execution sites, small mandatory installation requirements, and sessions
that outlive a particular browser connection. If deployment becomes
uniformly containerized or maintaining custom protocol code becomes more
expensive than installing mature libraries, the preferred choices may
change.

\hypertarget{three-connected-arguments}{%
\subsection{Three connected arguments}\label{three-connected-arguments}}

\textbf{Session assets motivate centralization.} Conversation histories,
summaries, memory, and tool traces can inform later work. Keeping their
management together makes them easier to inspect and associate across
execution sites. This is a practical choice for a self-hosted service;
distributed storage and federation are alternative architectures.
Child-side file journals remain on the child and are exposed through the
parent's interface. Centralization does not imply copying every artifact
to one filesystem, and it introduces availability, privacy, and
authority concentration costs.

\textbf{Low-dependency deployment makes projection practical.} Once
reasoning is centralized, a child used only for execution need not
independently invoke a model. With Python already present, the normal
runtime path can start without installing third-party Python packages.
Running the same code in a child role reduces the need for a separate
executor product. Each additional environment still requires
installation, credentials or tunnel admission, compatible tools, and
maintenance. We expect lower per-environment setup friction, not a
constant total cost as the fleet grows.

\textbf{Runtime composition limits special-purpose mechanisms.} Models,
tools, prompt templates, agents, skills, and environments are
addressable configuration data. Reusing their descriptions lets the
runtime change execution ownership without teaching the model a second
remote-only tool vocabulary. The debugger then adds an executor inside
the program that owns the evidence. Adapters and transport code remain
necessary; the reuse is at the abstraction and session-contract level.

These arguments were then probed at the mechanism level rather than left
as assertions. Routing a tool call to a child environment costs 1.8 to
2.2 network round-trips on the measured pair, and the parent's
forwarding layer, which attaches portable context and keeps journals,
adds 0.84 ms to a 4 KiB read and 12.19 ms to a 256 KiB write (Section
4). Centralization of reasoning and session state, by contrast, was not
compared against a decentralized alternative: it remains an assumption
this paper argues for rather than measures.

\begin{figure}[htbp]
\hypertarget{fig:rationale}{%
\centering
\includegraphics[width=\linewidth,keepaspectratio]{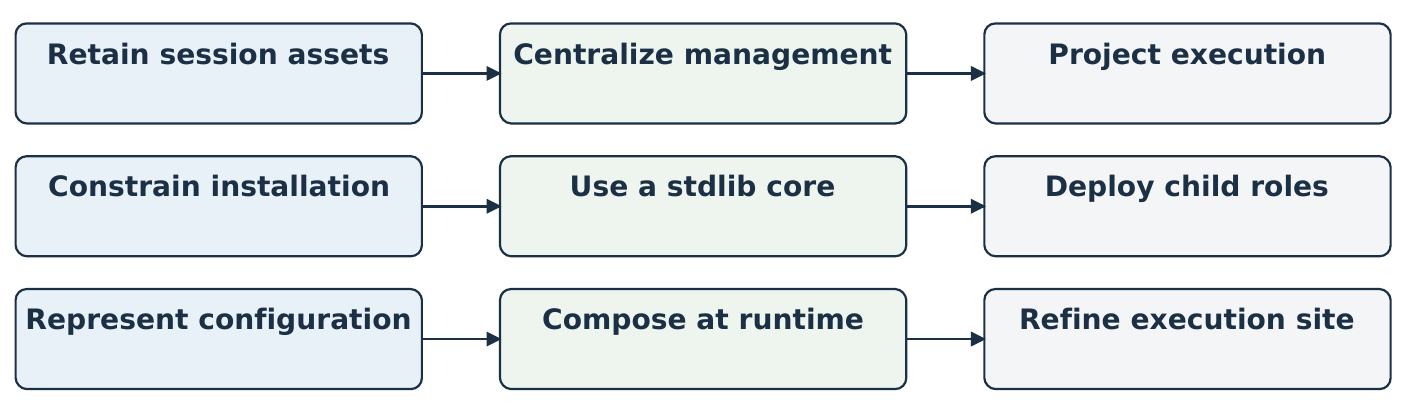}
\caption{Three design arguments. The arrows express conditional
engineering reasoning. Each expected benefit has deployment and
maintenance costs; none is a measured scaling law.}\label{fig:rationale}
}
\end{figure}

\hypertarget{expected-utility-and-recorded-tradeoffs}{%
\subsection{Expected utility and recorded
tradeoffs}\label{expected-utility-and-recorded-tradeoffs}}

We interpret each choice as an attempt to maximize expected lifecycle
value under current constraints and near-term assumptions:

\[
\begin{split}
U(d) = \mathbb{E}[&B_{\rm task}+B_{\rm reach}+B_{\rm evolution}\\
                 &-C_{\rm development}-C_{\rm deployment}\\
                 &-C_{\rm inference}-L_{\rm risk}].
\end{split}
\]

Here, \(d\) is a candidate design. The terms represent task benefit,
reachable-environment value, future adaptability, engineering and
deployment costs, inference cost, and risk loss. We have not estimated
their weights or probabilities. The expression disciplines the rationale
by requiring both benefit and cost to be stated; it does not prove that
the selected design is globally optimal. Table 1 condenses the master
draft's sixteen decision records into the choices central to this paper.

\begingroup
\small
\setlength{\tabcolsep}{4pt}
\begin{longtable}[]{@{}
  >{\raggedright\arraybackslash}p{(\columnwidth - 4\tabcolsep) * \real{0.28}}
  >{\raggedright\arraybackslash}p{(\columnwidth - 4\tabcolsep) * \real{0.36}}
  >{\raggedright\arraybackslash}p{(\columnwidth - 4\tabcolsep) * \real{0.36}}@{}}
\caption{Design choices as expected-value judgments. Benefits are
hypotheses supported by mechanisms, not comparative
measurements.}\tabularnewline
\toprule\noalign{}
\begin{minipage}[b]{\linewidth}\raggedright
Constraint and assumption
\end{minipage} & \begin{minipage}[b]{\linewidth}\raggedright
Choice and expected benefit
\end{minipage} & \begin{minipage}[b]{\linewidth}\raggedright
Cost and reconsideration condition
\end{minipage} \\
\midrule\noalign{}
\endfirsthead
\toprule\noalign{}
\begin{minipage}[b]{\linewidth}\raggedright
Constraint and assumption
\end{minipage} & \begin{minipage}[b]{\linewidth}\raggedright
Choice and expected benefit
\end{minipage} & \begin{minipage}[b]{\linewidth}\raggedright
Cost and reconsideration condition
\end{minipage} \\
\midrule\noalign{}
\endhead
\bottomrule\noalign{}
\endlastfoot
Package installation can be restricted; model SDKs change. &
Standard-library core and separate protocol adapters reduce mandatory
installation and provider coupling. & Custom HTTP, WebSocket, MCP, and
certificate code increases maintenance and audit work. Reconsider when
mature libraries offer greater net lifecycle value. \\
Session knowledge persists while execution sites vary. & Parent owns
inference and history; each request identifies one execution
environment. & Parent availability and network access become
dependencies. Federation or local autonomy may be preferable when those
dominate. \\
Model, tool, and prompt lifecycles differ from releases. & Registries
and runtime CRUD postpone binding and enable reconfiguration through the
service API. & Validation and consistency must handle invalid or
changing definitions. Static configurations remain useful for
reproducibility. \\
Some capabilities exist only at the caller. & Client-side descriptors
return tool calls to the session driver, avoiding an additional inbound
inspector. & Drivers must execute and return results correctly;
continuation, identity, and budgets become explicit obligations. \\
Context and inference have finite budgets. & Progressive skill
disclosure, independent subagent contexts, and a pre-send context guard
defer or compress information. & Selection and summarization can lose
information; token estimates are approximate. Larger windows do not
remove the need to measure these costs. \\
Networks permit different directions of reachability. & Direct HTTP and
child-initiated tunnels reuse application handlers; updates are pushed
over the available path. & Protocol, reconnect, proxy, and local replay
behavior require testing. Health of one path does not establish health
of another. \\
Broader reach increases the cost of mistakes. & Persist turns, record
execution sites, and expose reversible built-in file edits through
journals. & Journals do not undo arbitrary shell or external-system
effects. Enforceable tool admission remains a gap. \\
\end{longtable}
\endgroup

The cost side of that table is now partly measured, and two of its
implicit hypotheses failed in a specific way. Mechanisms that look like
alternate spellings of the same operation differ by more than an order
of magnitude depending on how completion is decided, not on where the
operation runs: a step in a persistent terminal costs about 50 ms where
a stateless call costs 3.7 to 4.1 ms, yet the terminal is the only arm
that completes tasks whose state lives in a process, and for a
pre-planable sequence a single scripted call beats both by one to two
orders of magnitude (29 ms against 573 ms for the same debugger
sequence). Changing only the completion strategy for the same commands
on the same arm changed cost by 4.2x to 19.6x while both arms completed
the task, and entering a layer whose prompt was not recognized cost
17.6x more than naming the prompt from the previous screen and 383x more
than the steady step on the same link (Table 3). What is still missing
is the benefit side: no arm in this program was model driven.

\hypertarget{composition-and-context-economics}{%
\subsection{Composition and context
economics}\label{composition-and-context-economics}}

The runtime represents models and tools as configuration objects and
manages conversation context separately. Four tool categories cover
in-process functions, MCP tools, skills, and client-side operations. A
skill is a disclosure item: its summary is initially visible and its
full instructions are inserted only when selected. A client-side tool
supplies a description without a server-side callable. Remote proxies
reuse ordinary descriptions but dispatch supported operations to a
child.

Multi-agent orchestration follows the same composition principle. An
orchestrator is an ordinary agent with a favorable position in the
task's information flow. It can invoke other agents as tools, with
separate models, tool sets, and contexts. Independent subagent contexts
can absorb large intermediate observations and return compressed
results. Compression can reduce what reaches the parent context, but may
discard important details; it does not eliminate task or budget limits.

Before a model request, the context guard estimates an input budget by
subtracting an output reserve from the configured window. It attempts
incremental compression, deeper summarization, and file spill in
sequence, retesting the estimate after each step. If the request still
exceeds the budget, it returns an error rather than sending a request
expected to be rejected. Unknown window size disables this guard. Spill
files must remain readable by the tools that will retrieve them, so
changing execution sites also requires attention to path visibility.
These mechanisms support the main architecture; their quality and
overhead have not been measured systematically.

\hypertarget{runtime-and-execution-contracts}{%
\section{Runtime and Execution
Contracts}\label{runtime-and-execution-contracts}}

\begin{figure}[htbp]
\hypertarget{fig:architecture}{%
\centering
\includegraphics[width=\linewidth,keepaspectratio]{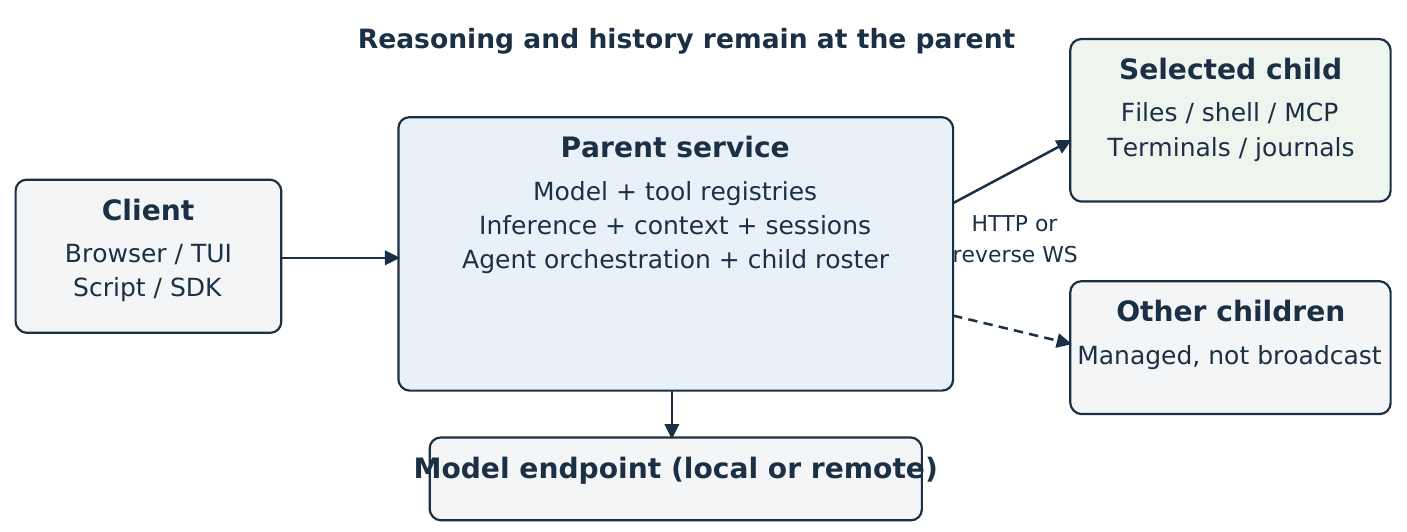}
\caption{Runtime ownership. Only one child is selected for a given
inference request. Parent-side orchestration can issue further model
requests while forwarding its execution tools to that child. The model
endpoint may be local or remote.}\label{fig:architecture}
}
\end{figure}

\hypertarget{service-structure}{%
\subsection{Service structure}\label{service-structure}}

agentsvc provides an HTTP service, a browser console, a terminal client,
and an embeddable Python runtime \citep{agentsvc}. Four upstream
adapters handle OpenAI Chat Completions, OpenAI Responses, native Ollama
chat, and Anthropic Messages. Model configuration selects an adapter at
runtime. The parent manages model and tool registries, sessions,
context, agent definitions, and a roster of child environments. A child
is another instance of the same service, with local tools, workspaces,
terminals, and journals (Figure \ref{fig:architecture}).

Configuration and session records use local files, including JSON
registries and per-session conversation, summary, and memory files. The
child's environment roster is local operational state: exporting or
pushing runtime configuration does not copy the parent's roster into the
child. This prevents an update from overwriting the managed topology
with another host's environment list.

The normal core path uses Python's standard library. The scanned
optional dependency allow-list contains \texttt{winpty} for Windows
pseudo-terminals. This scope excludes the browser build toolchain,
external MCP servers, optional speech-recognition components, and model
services. Python 3.7 is supported only for restricted local inference or
child-execution roles; full functionality requires Python 3.10 or later.
The compatibility statement is a declared product boundary, not a claim
that all operating-system and interpreter combinations have been tested.

\hypertarget{model-facing-descriptions-and-execution-ownership}{%
\subsection{Model-facing descriptions and execution
ownership}\label{model-facing-descriptions-and-execution-ownership}}

When a request selects a child, the model-facing catalogue comes from
the \textbf{parent's tool registry}. Names, parameter schemas, and
descriptions remain the same as in local mode. Per-request wrappers
forward execution operations using their original tool identifiers; they
do not register a second fleet of globally named remote tools. The
parent does not fetch the child's tool list on this main path
\citep{remote}.

This simplifies composition and prevents catalogue discovery from
depending on child availability. It also makes compatibility an
operational obligation: the child must implement the identifiers and
behavior described by the parent. Push updates and exported
configuration help maintain alignment, but there is no general
schema-negotiation or drift-proofing mechanism. A missing or
incompatible child tool fails when called. A disconnected child can
yield a tool error while the inference loop remains able to produce a
response; the task itself may still fail.

Execution ownership has explicit exceptions. \texttt{talk\_to} and
\texttt{delegate} stay at the parent because target-agent resolution and
orchestration state live there. Their subagents also reason at the
parent and use inherited remote wrappers for execution operations.
Skills remain disclosure items in the parent's loop, with full bodies
fetched from the child when required. Client-side tools follow their own
caller-owned handoff contract. Thus, selecting a child does not mean
blindly forwarding every model tool call. These ownership rules were
tested before this paper relied on them. One experiment ran a single
tool family through six routing arms: the same instance with and without
a forwarded request context, direct child calls with and without
context, a parent-forwarded direct HTTP call, and a parent-forwarded
reverse tunnel. Its fourteen tasks covered two-file journals with
repeated undo, over-undo, replace-all reversal, edit signatures, journal
turn-key collision, workspace pinning observed at three places, terminal
placement, directory visibility, dual-route reads of one workspace,
sessionless writes, catalogue comparison, and delegation. The reference
run recorded 168 task-runs across 84 condition-by-task cells and judged
them with 144 external checks: 142 passed, 2 failed, and none was
inconclusive. The two failures were the same cell in both batches, and
they are a statement about the contract rather than a flake: an arm that
carries no forwarded request context opened its terminal at the service
process's own working directory instead of the workspace, which is
exactly the kind of effect the context is supposed to carry.

\begingroup
\small
\setlength{\tabcolsep}{4pt}
\begin{longtable}[]{@{}
  >{\raggedright\arraybackslash}p{(\columnwidth - 6\tabcolsep) * \real{0.26}}
  >{\raggedright\arraybackslash}p{(\columnwidth - 6\tabcolsep) * \real{0.18}}
  >{\raggedright\arraybackslash}p{(\columnwidth - 6\tabcolsep) * \real{0.24}}
  >{\raggedright\arraybackslash}p{(\columnwidth - 6\tabcolsep) * \real{0.32}}@{}}
\caption{Tool semantics under six routing arms on one tool family,
judged by external effects rather than by model answers. Routing carries
bytes; the forwarded request context carries semantics. Pilot scale (two
repetitions per cell); this study owns the semantic result, while the
latency numbers quoted below come from the routing and terminal
studies.}\tabularnewline
\toprule\noalign{}
\begin{minipage}[b]{\linewidth}\raggedright
Routing arm
\end{minipage} & \begin{minipage}[b]{\linewidth}\raggedright
Forwarded request context
\end{minipage} & \begin{minipage}[b]{\linewidth}\raggedright
Journal directory
\end{minipage} & \begin{minipage}[b]{\linewidth}\raggedright
\texttt{undo} after two writes
\end{minipage} \\
\midrule\noalign{}
\endfirsthead
\toprule\noalign{}
\begin{minipage}[b]{\linewidth}\raggedright
Routing arm
\end{minipage} & \begin{minipage}[b]{\linewidth}\raggedright
Forwarded request context
\end{minipage} & \begin{minipage}[b]{\linewidth}\raggedright
Journal directory
\end{minipage} & \begin{minipage}[b]{\linewidth}\raggedright
\texttt{undo} after two writes
\end{minipage} \\
\midrule\noalign{}
\endhead
\bottomrule\noalign{}
\endlastfoot
Same instance, no context & absent & not created &
\texttt{NoSessionJournal}; 0 of 2 reverted \\
Same instance, with context & present & parent side, 2 snapshots &
\texttt{undone=true}; 2 of 2 \\
Direct child call, no context & absent & not created &
\texttt{NoSessionJournal}; 0 of 2 reverted \\
Direct child call, with context & present & child side &
\texttt{undone=true}; 2 of 2 \\
Parent-forwarded direct HTTP & present & child side &
\texttt{undone=true}; 2 of 2 \\
Parent-forwarded reverse tunnel & present & child side &
\texttt{undone=true}; 2 of 2 \\
\end{longtable}
\endgroup

Three consequences follow. First, the routing arm decided where bytes
went, while the forwarded context decided which side's journal
directory, terminal table, working directory, and search root a call
reached, and those effects landed at the executing side in every arm
that carried context. Second, the two transports are semantically
equivalent on this family: the error arrays recorded on the proxied and
tunnelled routes matched value by value, and journals appeared at the
executing side in both. A separate timing ladder, 24 timed calls per
condition, places the large step on the session-context axis rather than
the transport axis: 11.8 ms median for a context-free same-host call and
10.2 ms cross-host, against 138.9 ms for a parent-forwarded call
carrying context and 141.1 ms for the same call through the tunnel, a
2.2 ms transport difference. Third, the product contract is not
symmetric across arms in ways the interface announces. A child write
made without a session returns a reversible-looking handle
(\texttt{journal\_id=stateless/...}) although no journal directory
exists for it and \texttt{undo} answers \texttt{NoSessionJournal}. A
journal turn is keyed by the forwarded
\texttt{user\_message\_timestamp}, so two calls that reuse a one-second
timestamp merge into one turn and a single \texttt{undo} reverts both;
normal conversation rounds carry one timestamp each, so this bites
external callers rather than the browser. A forwarded workspace pin that
no longer resolves at the child falls back to the child's default
workspace after one warning line in the child log, which the caller
cannot see. And a declared \texttt{available\_tool\_ids} scope does not
gate the set of tools that can actually be called (Section 4).

A second group of measurements concerns how an interactive step is
judged complete. A persistent terminal is not a faster way to run a
command: on the same service a step in an open terminal cost 52.8 to
53.2 ms against 3.7 to 4.1 ms for a stateless call, and creating that
terminal cost about 358 ms locally and 266.95 ms at a child. What it
buys is capability. Driven one call per step, an agent failed every task
whose state lived inside a process (a read-evaluate loop, a debugger
session, an open database session, an input-dependent observation, and
working-directory and environment carry-over: 0 of 4 trials each) and
passed every task whose state could be externalized into a file, a
database, or a listening port. Two of those failures were silent: the
working-directory task wrote its file at the workspace root and the
environment task wrote an empty file, and no tool reported an error.

\begingroup
\small
\setlength{\tabcolsep}{4pt}
\begin{longtable}[]{@{}
  >{\raggedright\arraybackslash}p{(\columnwidth - 6\tabcolsep) * \real{0.34}}
  >{\raggedleft\arraybackslash}p{(\columnwidth - 6\tabcolsep) * \real{0.16}}
  >{\raggedleft\arraybackslash}p{(\columnwidth - 6\tabcolsep) * \real{0.16}}
  >{\raggedright\arraybackslash}p{(\columnwidth - 6\tabcolsep) * \real{0.34}}@{}}
\caption{Deciding that a step is finished is the largest interactive
cost, and it is not a property of the network path. Medians from a
deployed instance (three repetitions per cell) and a three-host
interactive chain (two runs of 28 trials; fast-cell medians agree to
within 7.4\%).}\tabularnewline
\toprule\noalign{}
\begin{minipage}[b]{\linewidth}\raggedright
Interaction situation
\end{minipage} & \begin{minipage}[b]{\linewidth}\raggedleft
Product default decision
\end{minipage} & \begin{minipage}[b]{\linewidth}\raggedleft
Explicit prompt pattern
\end{minipage} & \begin{minipage}[b]{\linewidth}\raggedright
Effect
\end{minipage} \\
\midrule\noalign{}
\endfirsthead
\toprule\noalign{}
\begin{minipage}[b]{\linewidth}\raggedright
Interaction situation
\end{minipage} & \begin{minipage}[b]{\linewidth}\raggedleft
Product default decision
\end{minipage} & \begin{minipage}[b]{\linewidth}\raggedleft
Explicit prompt pattern
\end{minipage} & \begin{minipage}[b]{\linewidth}\raggedright
Effect
\end{minipage} \\
\midrule\noalign{}
\endhead
\bottomrule\noalign{}
\endlastfoot
Seven-round interpreter session, whole session & 4,585.3 ms & 571.2 ms &
8.03x cheaper \\
Six-round debugger session & 2,524.4 ms & 518.9 ms & 4.86x \\
Four-round JavaScript session & 2,471.6 ms & 416.2 ms & 5.94x \\
One step inside a recognized interpreter & 51.8 to 53.3 ms & 51.8 to
53.3 ms & flat after the fix \\
Entering an unrecognized layer (20 entries, median) & 6,071.65 ms &
344.05 ms & 17.6x cheaper \\
Entering a layer behind an 8.2 s link establishment & 20,019.7 ms &
52.25 ms & 383x cheaper \\
Leaving a layer (exit transition, median) & 6,071.85 ms & 6,072.3 ms &
no arm is cheaper \\
Same traversal, end to end & 13,051.2 ms & 12,895.5 ms & 1.2\%
cheaper \\
\end{longtable}
\endgroup

\begin{figure}[htbp]
\hypertarget{fig:cost-ladder}{%
\centering
\includegraphics[width=\linewidth,keepaspectratio]{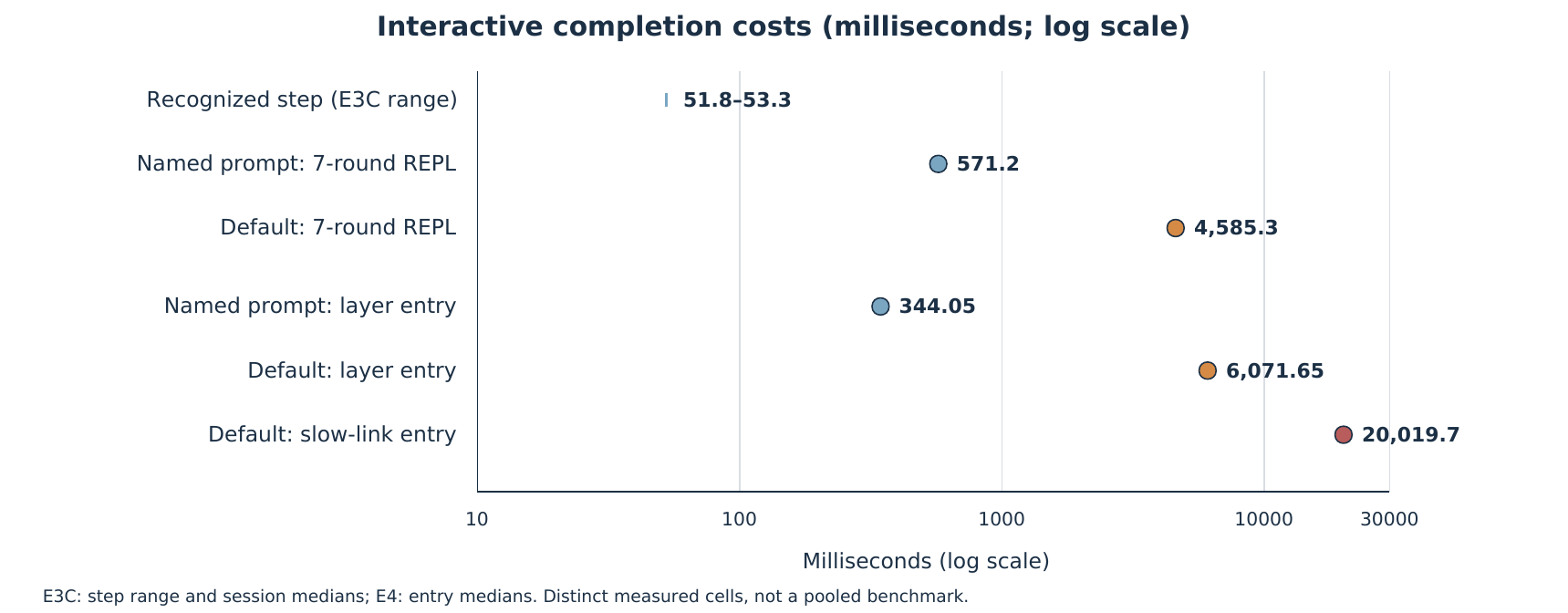}
\caption{Mechanism-level interactive costs on a logarithmic scale. The
marks reproduce the recognized-step range and selected medians from the
persistent-terminal and interactive-layer studies; they compare observed
cells rather than forming a unified throughput
benchmark.}\label{fig:cost-ladder}
}
\end{figure}

Completion decisions dominate. Changing only the strategy for deciding
that a step had finished, on the same arm with the same commands, moved
cost by factors of 4.2 to 19.6 while both strategies completed the task.
Naming the prompt pattern then collapsed the structure: 54 ordinary
steps across three interpreters and two strategies ran flat at 51.8 to
53.3 ms, and whole sessions fell by the factors in Table 3. An
unrecognized screen costs the caller's timeout plus a roughly constant
52.8 to 57.8 ms, so the worst case is bounded by the caller rather than
by the executor.

Crossing an interactive layer (a login to another host, a change of
identity, entering an interpreter) is a distinct case, because the
prompt that signals readiness is not known in advance. On a three-host
chain that included a hop whose link establishment alone took 8.2 s,
letting the product decide cost a median 6,071.65 ms per entering call
while taking the pattern from the previous screen's output cost 344.05
ms. Once inside a layer, the ordinary step was 51.85 to 53.4 ms on all
four arms and all seven chains, including the hop whose establishment
had just cost 8.2 s: path depth and operating system do not set the
per-step cost. Leaving a layer is the unhandled direction; the exit
transition cost about 6,072 ms on every arm, including the one that
names prompts, because the caller does not know the prompt of the layer
it returns to, which is why an entry 17.6x cheaper made the whole
traversal only 1.2\% cheaper. Two further observations are contract
issues rather than performance ones. Position is invisible from in-band
signals: on one nested chain all eight trials wrote their artifact on
the wrong host while every in-band check (the expected marker on screen,
correct artifact content, a healthy prompt) reported success, and
out-of-band checks put the artifact on the intended machine in 56 of 56
runs but on the intended host in only 48. The arm meant to catch this
with a one-line identity probe was defeated in an instructive way: the
probe was echoed, by the wrong interpreter. Identity changes are
likewise not tracked by tool-side state: after a change of user, writing
into a directory owned by the previous identity failed with
\texttt{Permission\ denied} in 51.8 ms while the prompt showed a normal
success state.

These interactive measurements are pilot scale (one to three repetitions
per cell, two independent runs for the layer study), on Linux hosts with
a fixed tool configuration and scripted drivers. They say nothing about
how a model chooses a strategy, and container and namespace entry could
not be exercised on the test hosts.

\hypertarget{client-side-handoff}{%
\subsection{Client-side handoff}\label{client-side-handoff}}

A tool with \texttt{tool\_type="client\_side"} describes an operation
that the service will never execute. The session driver supplies an
executor. Descriptors can be registered and selected by tool identifier,
or supplied through a request's \texttt{client\_tools} array. The latter
are session-scoped, absent from the global tool catalogue, and retained
in session metadata for subsequent requests \citep{clienttools}.
Duplicate names or identifiers are rejected.

When the model calls such a tool, the service finishes the current
inference request normally and returns the pending call. The driver
executes it locally and submits its result with the same session
identifier. The service can then continue reasoning with the result.
Built-in human questions, an explicitly enabled terminal-client local
shell, and debugger inspection use this mechanism.

Infrastructure remote execution and client-side handoff both separate
description from execution ownership. Their protocols differ. A remote
proxy obtains a child result \emph{within} the server-side tool loop
through HTTP or a tunnel. A client-side call ends the current request
and requires the caller to submit a continuation. This distinction
determines where budgets, retries, and failure handling belong.

\hypertarget{infrastructure-remote-execution}{%
\section{Infrastructure Remote
Execution}\label{infrastructure-remote-execution}}

\hypertarget{request-authority-and-durable-site-records}{%
\subsection{Request authority and durable site
records}\label{request-authority-and-durable-site-records}}

The parent manages multiple environments, but a request's
\texttt{remote\_env} selects one. It does not broadcast actions to the
fleet. A multi-target job must explicitly organize multiple requests or
sessions. For the selected request, execution operations consistently
address the child namespace; they do not silently alternate between
parent-local and child files.

\begin{figure}[htbp]
\hypertarget{fig:binding}{%
\centering
\includegraphics[width=\linewidth,keepaspectratio]{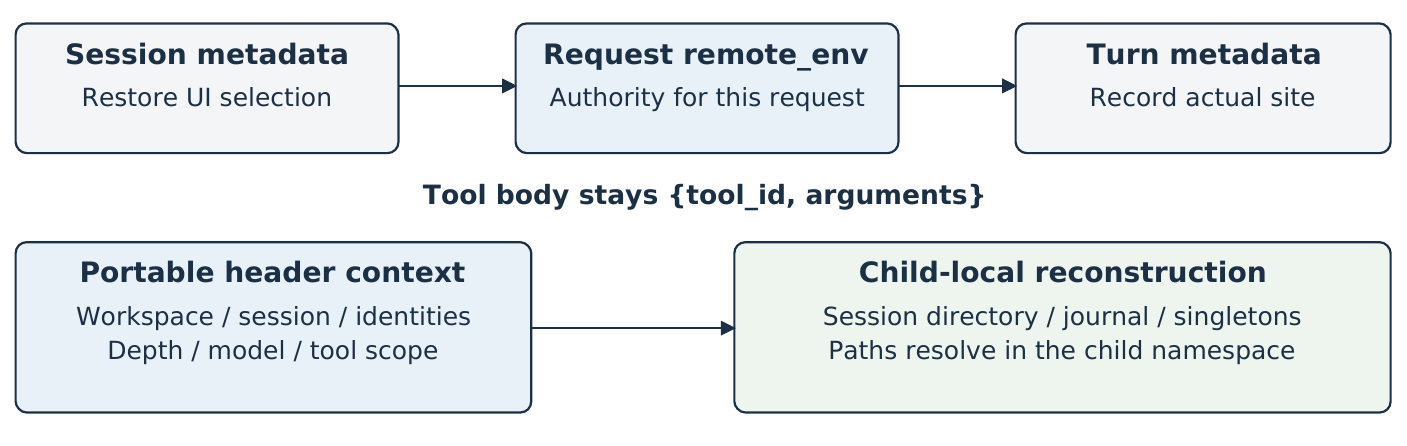}
\caption{Execution-site authority and context transfer. Session metadata
restores the client's selection; the current request controls routing;
turn metadata records what occurred. Portable values cross the boundary,
while host-bound objects are rebuilt at the child.}\label{fig:binding}
}
\end{figure}

Site selection has three different records (Figure \ref{fig:binding}).
The \textbf{request body} is authoritative for routing: an omitted
\texttt{remote\_env} selects local execution, while an explicitly
unknown environment is rejected. The \textbf{session metadata} persists
the UI's selected environment so that reopening a session can restore
the selection and resend it. The \textbf{turn metadata} records the
execution site used for that turn. Persisting a selection does not make
a later request inherit it automatically. API clients must supply it on
each inference request.

This distinction prevents historical metadata from being mistaken for
current routing authority. It also exposes a concrete integration
hazard: an API client that omits the field can run locally even when the
session previously used a child. The browser handles resubmission; other
clients must implement the contract. Recording the actual site supports
traceability, but cannot make arbitrary side effects reversible. Those
authority rules were probed rather than assumed. A bounded authorization
experiment ran a 124-row task family against isolated parent and child
instances and against the child reached through a deployed instance, in
two stages with the service authentication gate first off and then on:
authentication switch, forged request context, path and visibility
boundary, and landing enumeration for a synthetic secret. All 11 checks
passed, and the run archive verifies against 285 checksums.

\begingroup
\small
\setlength{\tabcolsep}{4pt}
\begin{longtable}[]{@{}
  >{\raggedright\arraybackslash}p{(\columnwidth - 4\tabcolsep) * \real{0.22}}
  >{\raggedright\arraybackslash}p{(\columnwidth - 4\tabcolsep) * \real{0.42}}
  >{\raggedright\arraybackslash}p{(\columnwidth - 4\tabcolsep) * \real{0.36}}@{}}
\caption{Authorization and request-context trust, 124 rows and 11 checks
on lab instances plus a deployed instance. A statement about deployment
state, not a security assessment.}\tabularnewline
\toprule\noalign{}
\begin{minipage}[b]{\linewidth}\raggedright
Control
\end{minipage} & \begin{minipage}[b]{\linewidth}\raggedright
Observed behavior
\end{minipage} & \begin{minipage}[b]{\linewidth}\raggedright
What it establishes
\end{minipage} \\
\midrule\noalign{}
\endfirsthead
\toprule\noalign{}
\begin{minipage}[b]{\linewidth}\raggedright
Control
\end{minipage} & \begin{minipage}[b]{\linewidth}\raggedright
Observed behavior
\end{minipage} & \begin{minipage}[b]{\linewidth}\raggedright
What it establishes
\end{minipage} \\
\midrule\noalign{}
\endhead
\bottomrule\noalign{}
\endlastfoot
Authentication gate off (default build) & The \texttt{/v1/} endpoints
used in this paper were reachable without credentials on the lab parent,
the lab child, and the deployed instance & Authorization is configured,
not inherited \\
Gate on & 401 on seven \texttt{/v1/} paths and on
\texttt{POST\ /v1/tools/call}; four credential forms accepted, a wrong
credential rejected & A working switch at the HTTP layer \\
Forged session identifier with a valid credential & Decided which
journal directory the write used & Context is not authenticated \\
Forged workspace pin with a valid credential & Decided where the working
directory and the write landed & Same \\
Recursion depth, agent identifiers, model identifier; unknown fields &
Accepted; unknown fields silently ignored, no schema rejection & The
contract is permissive \\
Declared \texttt{available\_tool\_ids} scope & Did not restrict the
callable set & Declared scope is not enforcement \\
Path boundary & Dot-dot traversal refused; an absolute path outside the
workspace written with status 200, after which a shell call read the
child's whole filesystem & The journal is recovery state, not an
isolation boundary \\
Synthetic secret, recursive search of the run tree & Written file 1,
child \texttt{chat\_data} 0, parent \texttt{chat\_data} 0, child stdout
log 0, the harness's own results row 1 & Landing enumeration only, not
an exfiltration study \\
\end{longtable}
\endgroup

Two conclusions follow, each with a hard edge. Authorization is a switch
and not a default: in the default build the endpoints used by this paper
were reachable without credentials, and with the gate enabled the
tool-call and registry paths returned 401 while all four credential
forms worked. That is not evidence of verified safety: WebSocket
attachment, terminal and prompt injection, skill and MCP surfaces, and
any tool that executes code were not exercised. The second conclusion
bears directly on the routing contract. After a valid credential is
presented, a forged request context still takes effect: an invented
session identifier decides where journals are written, an invented
workspace pin decides where a working directory resolves, and metadata
fields are accepted without schema validation. Request context is a
routing and reuse mechanism, not an authentication mechanism, and the
journal sits on the recovery side of that boundary rather than the
isolation side.

The catalogue boundary is thinner than it looks, though not in the way
one might expect. The lab parent and the lab child advertise identical
catalogues (12 tools in that harness configuration, same build and
registry), so drift is not what the measurements found. What they found
is scope mismatch and missing enforcement: the deployed instance
advertises 339 tools, including debugger- and evaluation-class entries
whose presence was measured and whose execution was not probed;
parent-side conversational tools are absent from the forwarded catalogue
on both proxy routes; a parent-only tool called while a remote
environment is bound returns an empty result rather than an error
(Section 4); and a declared \texttt{available\_tool\_ids} scope does not
gate what can be called. These measurements describe the default build
on lab instances and one dated snapshot of the deployed test instance;
they cannot be carried to a hardened or multi-tenant deployment, and the
deployed instance's authentication state must be re-probed before it is
cited.

\hypertarget{one-application-data-plane-two-transports}{%
\subsection{One application data plane, two
transports}\label{one-application-data-plane-two-transports}}

Both transports invoke the child's ordinary service interfaces. Tool
calls use \texttt{POST\ /v1/tools/call} with the body
\texttt{\{tool\_id,\ arguments\}}. Skill disclosure uses
\texttt{GET\ /v1/tools/skill/\{tool\_id\}}. Workspace access uses the
child's workspace endpoints. Direct mode sends these requests to a
reachable child. Tunnel mode sends their method, path, headers, and body
through a connection initiated by the child, which replays the request
against its own service \citep{remote}.

Portable request context travels in a base64url-encoded JSON value in
the \texttt{X-Agents-Request-Context} header. It includes workspace,
session identifier, user-message timestamp, recursion depth, agent
identities, model identifier, and available tool identifiers. The header
is an encoding and context-transfer convention, not encryption or
authorization. The body remains the ordinary tool-call contract; placing
workspace information in an extra body field does not substitute for the
context header.

The child reconstructs its own host-bound state, such as session
directories, journal holders, and server singletons. Matching session
identifiers associate child-side operations and journals with the parent
session without serializing Python objects or migrating conversation
management. A forwarded workspace must exist in the child's namespace.
If absent, the current implementation falls back to the child's default
workspace and records a warning. Consequently, sharing a path string is
not proof of sharing source files.

The browser reaches child workspaces through the parent's same-origin
environment bridge, keyed by a registered environment identifier. It
does not need to connect directly to children or hold their service
credentials. The bridge does not accept arbitrary destination URLs.
Terminals share one browser-to-parent WebSocket per tab; an open message
specifies the environment. Parent-to-child terminal bridges are created
lazily and released when the last terminal closes. Session inference
streams use SSE, separately from these terminal WebSockets. The cost of
this design was measured rather than asserted. A staged experiment
separated three things that are normally conflated: the transport
mechanism itself, per-call HTTP routing to an already deployed child,
and a child-initiated reverse tunnel to that same child. Every arm had
to prove its route before being timed, by showing that a value held only
on the child could be obtained only through the measured route and that
a parent-side witness was not readable at the child. Three runs for the
routing stage and four for the tunnel stage were merged on that basis,
giving n=63 per cell.

\begingroup
\small
\setlength{\tabcolsep}{4pt}
\begin{longtable}[]{@{}
  >{\raggedright\arraybackslash}p{(\columnwidth - 10\tabcolsep) * \real{0.16}}
  >{\raggedright\arraybackslash}p{(\columnwidth - 10\tabcolsep) * \real{0.1}}
  >{\raggedleft\arraybackslash}p{(\columnwidth - 10\tabcolsep) * \real{0.11}}
  >{\raggedleft\arraybackslash}p{(\columnwidth - 10\tabcolsep) * \real{0.2}}
  >{\raggedleft\arraybackslash}p{(\columnwidth - 10\tabcolsep) * \real{0.22}}
  >{\raggedleft\arraybackslash}p{(\columnwidth - 10\tabcolsep) * \real{0.21}}@{}}
\caption{Median milliseconds per operation at the routing stage: the lab
parent and a deployed child on another host, 3 pooled runs, 63 samples
per cell. Routing one call costs 1.8 to 2.2 round-trips, and the
context-attaching, journal-keeping forwarding layer adds 0.84 to 12.19
ms.}\tabularnewline
\toprule\noalign{}
\begin{minipage}[b]{\linewidth}\raggedright
Operation
\end{minipage} & \begin{minipage}[b]{\linewidth}\raggedright
Payload
\end{minipage} & \begin{minipage}[b]{\linewidth}\raggedleft
Local
\end{minipage} & \begin{minipage}[b]{\linewidth}\raggedleft
Direct HTTP to child
\end{minipage} & \begin{minipage}[b]{\linewidth}\raggedleft
Parent-forwarded
\end{minipage} & \begin{minipage}[b]{\linewidth}\raggedleft
Forwarding cost
\end{minipage} \\
\midrule\noalign{}
\endfirsthead
\toprule\noalign{}
\begin{minipage}[b]{\linewidth}\raggedright
Operation
\end{minipage} & \begin{minipage}[b]{\linewidth}\raggedright
Payload
\end{minipage} & \begin{minipage}[b]{\linewidth}\raggedleft
Local
\end{minipage} & \begin{minipage}[b]{\linewidth}\raggedleft
Direct HTTP to child
\end{minipage} & \begin{minipage}[b]{\linewidth}\raggedleft
Parent-forwarded
\end{minipage} & \begin{minipage}[b]{\linewidth}\raggedleft
Forwarding cost
\end{minipage} \\
\midrule\noalign{}
\endhead
\bottomrule\noalign{}
\endlastfoot
read & 4 KiB & 2.29 & 16.56 & 17.40 & +0.84 \\
read & 64 KiB & 3.08 & 35.70 & 36.84 & +1.14 \\
read & 256 KiB & 4.81 & 447.50 & 451.35 & +3.85 \\
write & 4 KiB & 2.58 & 18.04 & 20.46 & +2.42 \\
write & 64 KiB & 2.90 & 36.22 & 40.65 & +4.43 \\
write & 256 KiB & 3.74 & 59.06 & 71.25 & +12.19 \\
\end{longtable}
\endgroup

Three readings matter for the architecture claim. A remote call sits
close to the network floor: measured round-trip time on that pair was
6.68 to 8.47 ms at the minimum and 7.48 to 13.35 ms at the median, a 4
KiB call to the child costs 14.3 ms over the local control, about 1.9
round-trips on top of a 2.3 ms local service cost, and the same byte
counts came back on every arm, which argues against a hidden
serialization layer. On the tunnel path the 4 KiB read sits about 5.5 ms
above the round-trip floor (13.24 ms against a 7.79 ms in-run median),
which is where the 1.8-round-trip figure comes from. The parent's richer
path is cheap on reads and grows on writes, so forwarding large writes
is a real budget item even though the layer is not expensive. And the
reverse tunnel is not the expensive option; on that pair it was the only
available one, because the child's own HTTP endpoint refused connections
arriving from the parent's network. Reachability asymmetry is a measured
property of this environment, not an assumption. Reuse also pays under
load: against a local arm that dials one connection per call, the
cross-host tunnel is worse at n=1 (12.92 against 2.63 ms), level at n=8,
and lower at n=16 (38.01 against 44.03 ms) in four of four runs, because
in-flight multiplexing avoids per-call connection setup.

\begingroup
\small
\setlength{\tabcolsep}{4pt}
\begin{longtable}[]{@{}
  >{\raggedright\arraybackslash}p{(\columnwidth - 6\tabcolsep) * \real{0.32}}
  >{\raggedleft\arraybackslash}p{(\columnwidth - 6\tabcolsep) * \real{0.12}}
  >{\raggedleft\arraybackslash}p{(\columnwidth - 6\tabcolsep) * \real{0.12}}
  >{\raggedright\arraybackslash}p{(\columnwidth - 6\tabcolsep) * \real{0.44}}@{}}
\caption{A reused reverse tunnel against a fresh HTTP call per
operation. Tunnel figures come from a lab child dialing the parent, HTTP
figures from the deployed child on the same host pair; 4 and 3 pooled
runs, 12 and 63 samples per cell. The 64 KiB crossover is reported as
measured rather than smoothed. A throughput figure quoted from one
connection mode is not a property of the transport: the same byte volume
ran at 0.59 MB/s per call and 2.9 MB/s on the reused
tunnel.}\tabularnewline
\toprule\noalign{}
\begin{minipage}[b]{\linewidth}\raggedright
Operation and payload
\end{minipage} & \begin{minipage}[b]{\linewidth}\raggedleft
Reverse tunnel
\end{minipage} & \begin{minipage}[b]{\linewidth}\raggedleft
Per-call HTTP
\end{minipage} & \begin{minipage}[b]{\linewidth}\raggedright
Reading
\end{minipage} \\
\midrule\noalign{}
\endfirsthead
\toprule\noalign{}
\begin{minipage}[b]{\linewidth}\raggedright
Operation and payload
\end{minipage} & \begin{minipage}[b]{\linewidth}\raggedleft
Reverse tunnel
\end{minipage} & \begin{minipage}[b]{\linewidth}\raggedleft
Per-call HTTP
\end{minipage} & \begin{minipage}[b]{\linewidth}\raggedright
Reading
\end{minipage} \\
\midrule\noalign{}
\endhead
\bottomrule\noalign{}
\endlastfoot
4 KiB read & 13.24 & 16.56 & tunnel cheaper by 3.3 ms \\
64 KiB read & 41.60 & 35.70 & tunnel slower by 5.9 ms \\
256 KiB read & 91.81 & 447.50 & tunnel 4.9x faster \\
256 KiB write & 49.27 & 59.06 & tunnel slightly faster \\
4 KiB read at n=1 / 8 / 16: tunnel versus the local arm dialing one
connection per call & 12.92 / 27.96 / 38.01 & 2.63 / 28.32 / 44.03 &
tunnel worse at n=1, level at n=8, better at n=16 (4 of 4 runs) \\
4 KiB read at n=1 / 8 / 16: per-call HTTP to the child, and the same
calls proxied by the parent & 16.29 / 21.35 / 22.86 (child) & 16.58 /
24.75 / 23.27 (proxied) & proxying costs 0.3, 3.4, and 0.4 ms \\
\end{longtable}
\endgroup

Two findings made while validating transport belong to this section
because they changed the product. A Nagle and delayed-ACK interaction
had inflated that same tunnel path to 97.24 and 93.68 ms in two runs;
instrumented with \texttt{TCP\_NODELAY} the same path measured 10.34 and
10.95 ms, and a bare four-frame probe measured 44.02 ms with Nagle
against 0.13 ms for a single write and 0.14 ms with that option set. The
defect was located at the mediation layer and is fixed in the current
code, and every routing and tunnel number quoted above was measured
after that fix. The second is a measurement hazard rather than a product
defect: the same local read was about 25x slower when the run root sat
on a Windows-mounted path, which is why run roots are required to sit on
native Linux storage. Both are the same kind of result: what looked like
a property of remote execution was a property of one layer underneath
it.

\hypertarget{reverse-tunnel-lifecycle-and-framing}{%
\subsection{Reverse tunnel lifecycle and
framing}\label{reverse-tunnel-lifecycle-and-framing}}

\begin{figure}[htbp]
\hypertarget{fig:tunnel}{%
\centering
\includegraphics[width=\linewidth,keepaspectratio]{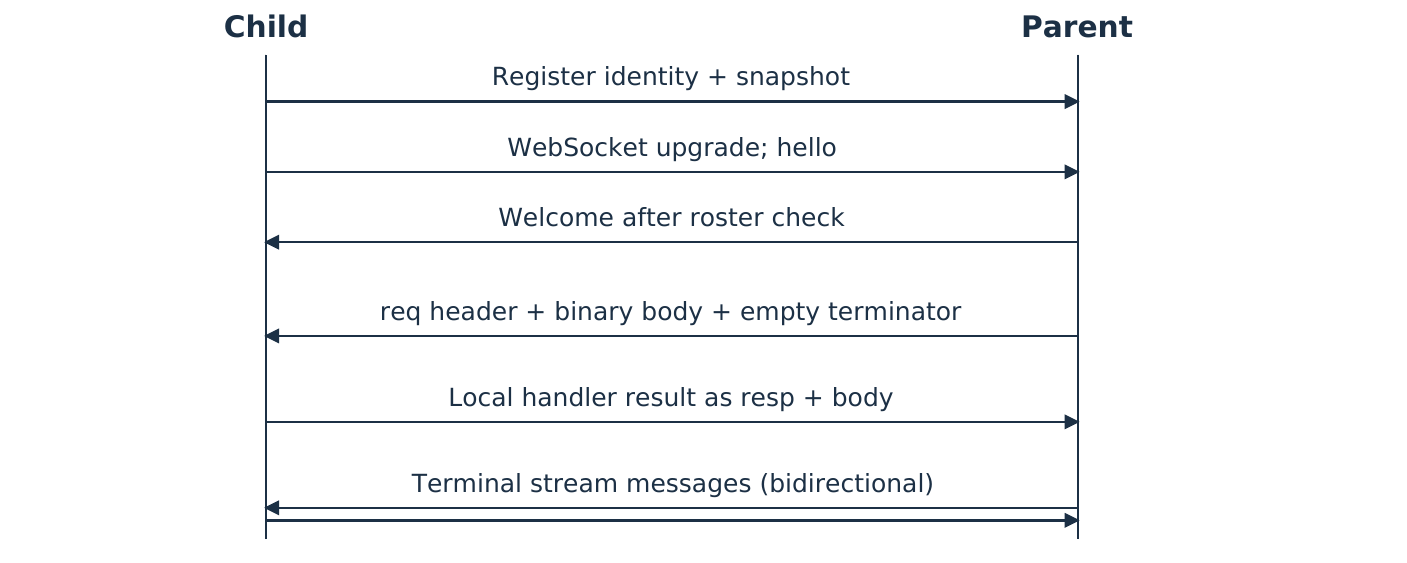}
\caption{Reverse tunnel lifecycle and data path. Admission and
connection ownership belong to the control plane. Requests and terminal
streams share the established connection but execute through the child's
local service.}\label{fig:tunnel}
}
\end{figure}

A child first registers its persisted tunnel identity and version
snapshot with the parent, then initiates a WebSocket upgrade and sends
\texttt{hello}. The parent checks its roster and returns
\texttt{welcome}. An unregistered identity is rejected. A replacement
connection for the same identity displaces the old connection; deleting
the environment sends \texttt{deregister} so that the child disables
reconnection (Figure \ref{fig:tunnel}). Registration is a separate
child-initiated HTTP exchange before the persistent data connection.

HTTP request and response messages carry JSON headers followed by binary
body frames of up to 256 KiB. An empty binary frame terminates a body,
allowing streaming without a predeclared body length. Each tunnel admits
up to 16 in-flight HTTP calls; the child uses eight local request
workers. Bodies up to 8 MiB remain in memory and larger bodies spill to
a temporary file. These are engineering defaults, not experimentally
optimized parameters.

Terminal streams use a separate message family: \texttt{stream-open},
\texttt{stream-ready} or \texttt{stream-error}, bidirectional
\texttt{stream-data}, and \texttt{stream-close}. Stream data is
base64-encoded. The parent sends WebSocket protocol-level pings every 30
seconds and disconnects after 90 seconds without receiving a frame. The
child read timeout is 95 seconds and reconnect backoff ranges from 1 to
30 seconds. Ordinary tunneled HTTP calls default to a 600-second
forwarding timeout; update pushes use 900 seconds. These transport
limits must be distinguished from tool-specific and model-request
limits.

Outbound WebSocket dialing supports HTTP CONNECT proxies and
\texttt{HTTP\_PROXY}, \texttt{HTTPS\_PROXY}, and \texttt{NO\_PROXY},
including host, IP, CIDR, and port rules. CONNECT creates a byte tunnel.
Only correctly validated end-to-end TLS protects the subsequent
WebSocket handshake path and query from an ordinary forwarding proxy;
plaintext WebSockets do not provide that confidentiality. A
TLS-terminating reverse proxy can see decrypted requests and must
control credential logging separately.

\hypertarget{local-replay-credentials-and-failure-semantics}{%
\subsection{Local replay, credentials, and failure
semantics}\label{local-replay-credentials-and-failure-semantics}}

The child's replay target is derived from its actual listening address,
port, and protocol. Wildcard IPv4 and IPv6 bindings map to their
loopback equivalents; a concrete interface binding uses that address.
HTTPS children use TLS for local HTTP replay and \texttt{wss} for
terminal bridging. Both request and stream paths share target
resolution. This fixes the earlier failure in which registration
succeeded while plaintext replay against an HTTPS child failed.

When child authentication is enabled, the child issues its own local
session credential for HTTP replay and a short-lived setup token for
stream handshakes. In reverse-tunnel mode, the parent therefore does not
need to store the child's service credential. Direct mode does require
the parent to obtain or hold child credentials. Neither design removes
the parent's authority to operate connected children.

Local replay TLS disables certificate verification by default to
accommodate locally generated certificates.
\texttt{AGENTS\_TUNNEL\_LOOPBACK\_VERIFY=1} enables the default
verification context, requiring an appropriately trusted certificate and
matching identity. The default is a security tradeoff under a
trusted-local-host assumption. Tunnel admission does not
cryptographically authenticate an unverified local TLS peer,
particularly if replay uses a concrete interface address. Certificate
pinning is a possible future strengthening.

The local connection timeout is 10 seconds. After connecting, the socket
timeout is removed so large uploads and streamed responses are not
truncated by that connect limit. This bounds connection establishment,
not the duration of an established request: a stalled response can still
retain a child worker. Parent forwarding timeouts do not by themselves
establish cancellation of the child's operation.

A local \texttt{OSError} becomes an HTTP 502 response containing the
resolved target; other unexpected local-call exceptions become 500
responses. The parent formats these failures as model-visible tool
errors instead of relying solely on a forwarding timeout. Registration
and online status do not probe local replay, so a live control
connection can coexist with a broken data plane. No degraded health
state currently represents that distinction. Failure semantics were
probed by injecting faults into the deployed instance rather than by
reading configuration: nine injected conditions against a ten-task
matrix, 264 rows in all (151 clean, 47 marked invalid by the harness
rule, 61 error), with all 11 failing cells explained, followed by stages
that isolate the pseudo-terminal path, the cause of terminal
degradation, and transport-level faults. A network-layer partition could
not be constructed, because neither host had \texttt{ip},
\texttt{iptables}, \texttt{nft}, or \texttt{tc}; the transport faults
are covered from the observable sides instead.

\begingroup
\small
\setlength{\tabcolsep}{4pt}
\begin{longtable}[]{@{}
  >{\raggedright\arraybackslash}p{(\columnwidth - 4\tabcolsep) * \real{0.24}}
  >{\raggedright\arraybackslash}p{(\columnwidth - 4\tabcolsep) * \real{0.4}}
  >{\raggedright\arraybackslash}p{(\columnwidth - 4\tabcolsep) * \real{0.36}}@{}}
\caption{Fault-injection semantics on a deployed instance: 9 conditions
by 10 tasks, 264 rows, plus three follow-up stages. Mechanism level
only; no model took part.}\tabularnewline
\toprule\noalign{}
\begin{minipage}[b]{\linewidth}\raggedright
Injected condition
\end{minipage} & \begin{minipage}[b]{\linewidth}\raggedright
What the deployed service did
\end{minipage} & \begin{minipage}[b]{\linewidth}\raggedright
Consequence for the contract
\end{minipage} \\
\midrule\noalign{}
\endfirsthead
\toprule\noalign{}
\begin{minipage}[b]{\linewidth}\raggedright
Injected condition
\end{minipage} & \begin{minipage}[b]{\linewidth}\raggedright
What the deployed service did
\end{minipage} & \begin{minipage}[b]{\linewidth}\raggedright
Consequence for the contract
\end{minipage} \\
\midrule\noalign{}
\endhead
\bottomrule\noalign{}
\endlastfoot
Command longer than the call timeout & Returned at 3,025.5 and 3,026.0
ms with \texttt{status\ 200}; the file the command was to create
appeared out-of-band, and a residual process was still running & A
timeout ends the call, not the work \\
Client reset 1, 3, and 10 s into a 12 s command & Work not cancelled in
any repetition; the first call after the reset queued behind the orphan
for 7,047.8, 5,049.6, and 76.2 ms & One terminal is a queue; an
abandoned call delays the next one \\
12 concurrent commands on one terminal & All 12 returned 200, serialized
over 24.19 s of wall time (median 14,112.3 ms, maximum 24,181.0 ms,
spacing 2.01 to 2.02 s) with zero queue errors & Good isolation, but a
terminal is not a concurrency domain \\
Refused versus unreachable dependency & A refusal came back in about 0.1
ms; a black-hole address consumed the whole 5 s budget (5.0 and 5.01 s)
& Unreachability is bounded by whichever client dials it, not by the
caller's timeout \\
1 MiB command argument & Stateless path: \texttt{status\ 400},
\texttt{Argument\ list\ too\ long}, 135.0 ms. Terminal path: executed in
3,427.9 ms & The same tool has different input limits on each path \\
200k lines of output & 1,288,894 bytes returned in 1,070.5 and 2,149.9
ms, \texttt{truncated=False}, byte-identical between repetitions,
service unchanged at 26 file descriptors and 26 threads & Large reads
are transferred rather than truncated \\
16 concurrent 20 s data commands & Control plane unaffected: health
endpoint at 16.3 and 16.8 ms median over 175 probes (95th percentile
20.3 and 22.0, maximum 38.1 and 48.6), and a new terminal was still
admitted in 18.7 and 18.4 ms & Data-plane load does not crowd out the
control plane \\
A 65th call with 17 sessions registered and \texttt{max\_terminals\ 64}
& Accepted in 17.3 and 18.9 ms, with no explicit rejection and no silent
failure & Capacity presents as eviction of least-recently-used idle
terminals (120 s grace), not as an error \\
1,021 idle connections from one client & 912 threads and 909 file
descriptors, about 89\% of the 1,024 soft limit, after which the control
plane hit a 5 s timeout & Tenant boundaries are weak \\
Tool visibility during a fault & A 344-tool catalogue was served in
148.8 and 1,784.3 ms while the execution tool was missing from the
agent's own tool table and the health endpoint answered 200 & Catalogue
availability and control-plane health are different signals \\
\end{longtable}
\endgroup

The headline result is a mode fork. The same command, issued from three
session arms of one run, behaved as two different products. A session
that owned a pseudo-terminal answered its first call in 281.6 and 280.6
ms, honored a supplied timeout (returning an empty result at 1,271.9 and
1,271.1 ms), and kept the working directory across calls. The stateless
arms answered in 16.6 to 20.9 ms, ignored the same timeout and returned
at 5,021.4 to 5,022.8 ms, and reset the working directory. Nothing in
the response says which path ran: there is no mode field, no registry
entry for a stateless call, and in the service's 7,647,551-byte
INFO-level log the fallback message occurs zero times, because a
terminal-creation failure is caught and logged at debug level before the
code runs a subprocess instead. A separate probe of 40 fresh sessions
made the fork visible as a count: the first ten calls received a
terminal (257.1 to 259.2 ms) and the remaining thirty answered in 2.8 to
4.1 ms with the registry frozen at ten.

Degradation was not resource exhaustion. Opening 200 pseudo-terminals
from inside the service moved the kernel counter from 41 to 241 against
a 4,096 maximum; file descriptors, threads, process counts, available
memory, queue depth, request shape, and container pseudo-terminal
capacity were each measured and ruled out. What was measured is an
admission count that resets when terminals are deleted and does not heal
by itself (eight probes over 105 s, all stateless), and that drifts: the
pseudo-terminal share of calls moved from 72 of 72, to 36 of 260, to 0
of 72, to 10 of 40, to 10 of 30 and then 0 of 12 across one evening of
runs. The trigger is not attributed to any named constant, environment
override, or operating-system resource, and this paper does not name
one; the mechanism claim is the silent two-path degradation and its
invisibility to the caller and to the log.

Two further results matter for operation. A fault's cost depends on its
kind: a refusal is nearly free, an unreachable target consumes whatever
budget the dialing client holds, and the stateless fallback runs with no
internal timeout at all, so the caller's timeout does not bound it. And
the boundary between tenants is weak in one direction and strong in
another: sixteen concurrent long data commands did not disturb the
control plane, while a single client holding 1,021 idle connections
pushed the service to 912 threads and 909 file descriptors and made that
same control plane time out.

\hypertarget{deployment-and-operations}{%
\subsection{Deployment and operations}\label{deployment-and-operations}}

\begin{figure}[htbp]
\hypertarget{fig:topologies}{%
\centering
\includegraphics[width=\linewidth,keepaspectratio]{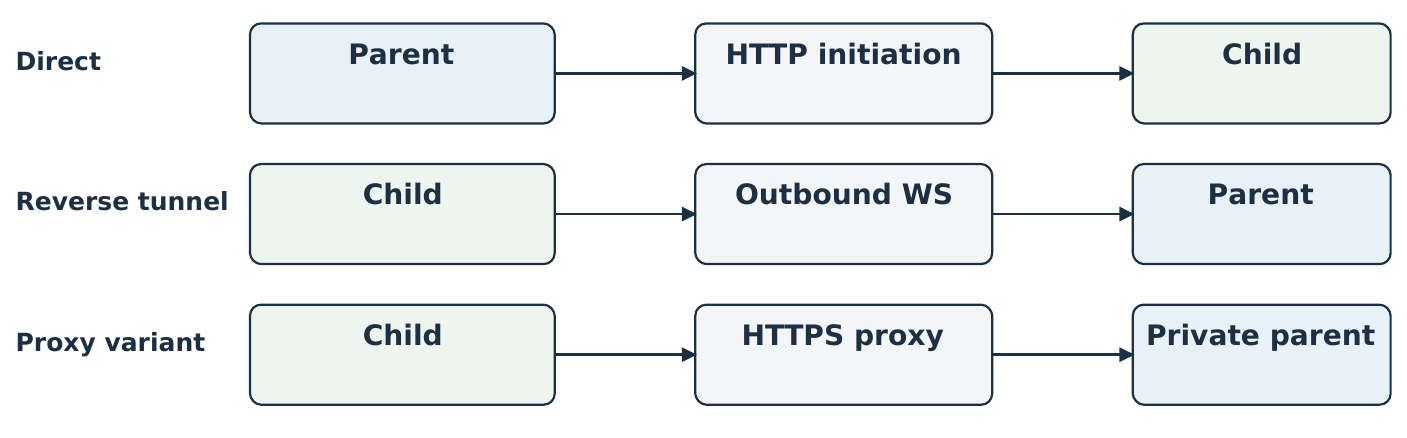}
\caption{Three reachability patterns. Arrows show connection initiation.
The HTTPS reverse proxy must reach the parent and be its only
untrusted-network ingress when the parent relies on proxy
authentication. An outbound CONNECT proxy is a separate, optional
hop.}\label{fig:topologies}
}
\end{figure}

The implementation supports three documented reachability patterns
(Figure \ref{fig:topologies}). Direct mode requires parent-to-child
reachability; the child need not initiate a connection to the parent.
Reverse-tunnel mode requires child-to-parent reachability, allowing a
child behind NAT to participate. A variant places the parent behind an
authenticated public HTTPS reverse proxy while the child connects
outward to that proxy. The last arrangement assumes an actual
proxy-to-parent route; it does not connect two unreachable private
networks without additional networking.

The parent checks child version and inference-status snapshots and
pushes incremental archives over the same application path. Direct
updates consequently need no child-to-parent connectivity; tunnel
updates reuse the child-initiated connection. An installation endpoint
exports a self-extracting package containing runtime code, a built
frontend, and selected configuration. Initial remote access and
deployment permission must already exist. These mechanisms reduce
repeated deployment work but do not demonstrate fleet-scale operational
cost or reliable self-replication.

The service authentication gate covers tunnel and environment-bridge
endpoints. If it is disabled, any client that can reach the parent can
use its exposed authority, including tunnel registration and child
access. A parent relying on reverse-proxy authentication must be
unreachable through an unprotected alternative route. Service-level
admission also does not provide per-tool permissions, a sandbox, or
multi-tenant isolation.

\hypertarget{execution-site-switching-diagnose-remotely-repair-locally}{%
\subsection{Execution-site switching: diagnose remotely, repair
locally}\label{execution-site-switching-diagnose-remotely-repair-locally}}

Remote execution is usually presented as running one tool somewhere
else. The stronger claim is that the site can be switched inside one
session and switched back: a diagnosis performed where the failure lives
is repaired where the source lives, in the same conversation. The switch
is a property of the inference request rather than of each tool call.
While \texttt{remote\_env} is bound, every tool offered in that round
executes in the registered environment, and the turn metadata records
the site actually used; model inference and conversation history remain
on the parent. The diagnosis therefore returns as ordinary tool results
inside parent-side history, not through a copy step. We measured this
loop with a deterministic scripted model endpoint and out-of-band
assertions against the two service data directories (run
\texttt{e9-20261124-231824}, protocol \texttt{e9-envswitch-v1c}: 16
checks, all passing; four repetitions for timing, one for each semantic
probe).

\begin{figure}[htbp]
\hypertarget{fig:e9-timeline}{%
\centering
\includegraphics[width=\linewidth,keepaspectratio]{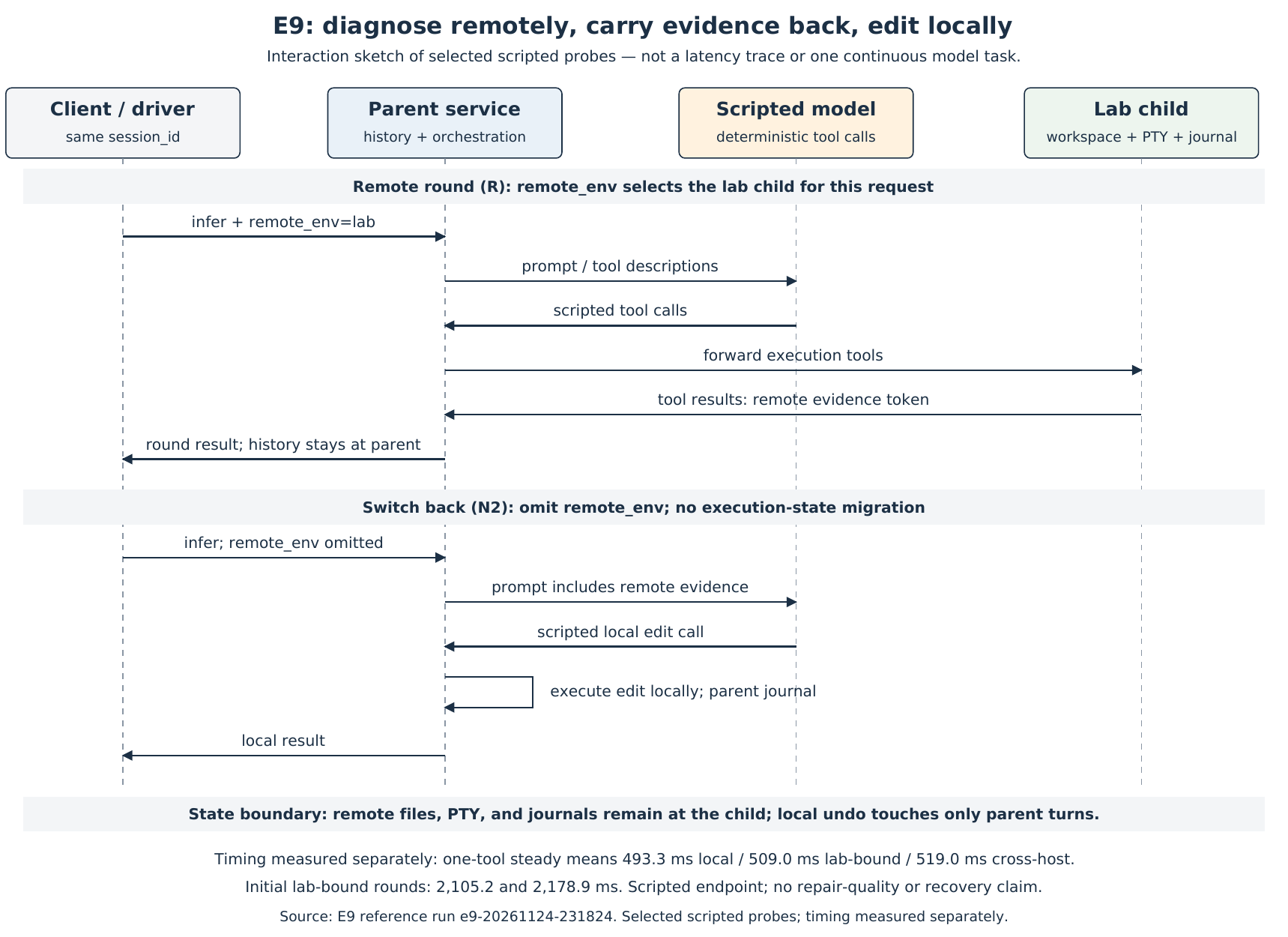}
\caption{Execution-site switching within one parent-side conversation. A
bound request forwards execution tools to the child and returns evidence
to the parent history; a subsequent local request can use that evidence
while editing locally. The sequence illustrates selected scripted probes
rather than a continuous model task or latency trace. Files, terminals,
and journals remain at their execution sites.}\label{fig:e9-timeline}
}
\end{figure}

Figure \ref{fig:e9-timeline} illustrates the request and state
boundaries in the scripted switching probes. The browser controls for
choosing the execution site are shown immediately afterward.

\begin{figure}[htbp]
\hypertarget{fig:web-ui}{%
\centering
\includegraphics[width=\linewidth,keepaspectratio]{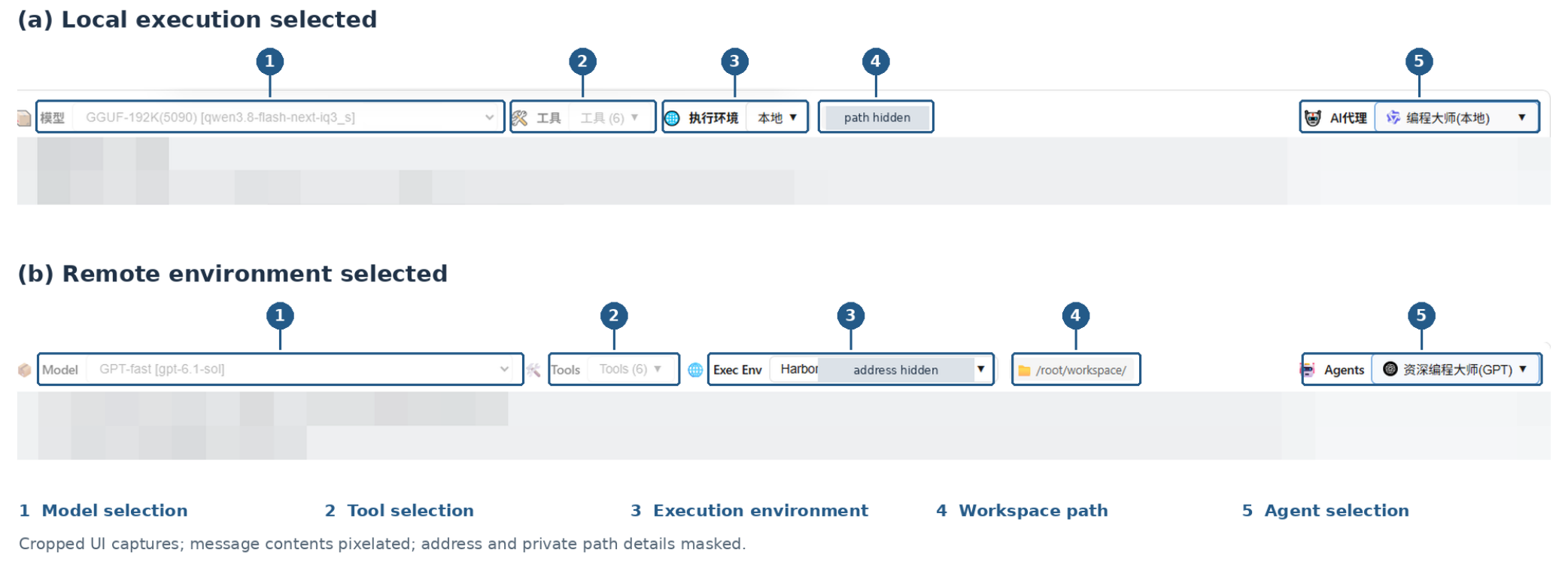}
\caption{Browser controls for execution-site selection: (a) local
execution and (b) a selected remote environment. Numbered annotations
identify model selection (1), tool selection (2), execution environment
(3), workspace path (4), and agent selection (5). The captures are
cropped; conversation contents are pixelated, and the remote address and
private project path are masked.}\label{fig:web-ui}
}
\end{figure}

Figure \ref{fig:web-ui} shows the browser controls for local and remote
execution: model selection, tool selection, execution environment,
workspace path, and agent selection. The two captures show different
sessions and model configurations; they illustrate the interface rather
than a controlled execution-equivalence test. Terminal and file-manager
panels are not expanded in these views. The switching mechanism behind
this control, and what it costs, is measured in this section.

\begingroup
\small
\setlength{\tabcolsep}{4pt}
\begin{longtable}[]{@{}
  >{\raggedright\arraybackslash}p{(\columnwidth - 4\tabcolsep) * \real{0.21}}
  >{\raggedright\arraybackslash}p{(\columnwidth - 4\tabcolsep) * \real{0.44}}
  >{\raggedright\arraybackslash}p{(\columnwidth - 4\tabcolsep) * \real{0.35}}@{}}
\caption{One-session execution-site switching with a scripted model
endpoint. Timing repetitions are small (n=4 per condition, single
machine, no concurrency), the cross-host case uses one deployed
endpoint, and the measured switch is the request field, not the browser
selector in Figure \ref{fig:web-ui}.}\tabularnewline
\toprule\noalign{}
\begin{minipage}[b]{\linewidth}\raggedright
Observation inside one session
\end{minipage} & \begin{minipage}[b]{\linewidth}\raggedright
Measured result
\end{minipage} & \begin{minipage}[b]{\linewidth}\raggedright
Consequence for the workflow
\end{minipage} \\
\midrule\noalign{}
\endfirsthead
\toprule\noalign{}
\begin{minipage}[b]{\linewidth}\raggedright
Observation inside one session
\end{minipage} & \begin{minipage}[b]{\linewidth}\raggedright
Measured result
\end{minipage} & \begin{minipage}[b]{\linewidth}\raggedright
Consequence for the workflow
\end{minipage} \\
\midrule\noalign{}
\endhead
\bottomrule\noalign{}
\endlastfoot
Site proof across local, bound same-host, bound cross-host, and switched
back & Each round verified by a site-specific nonce or machine ID; no
round executed at the wrong site & Site selection is per request round,
reversible, and verifiable after the fact \\
Diagnosis carryover & All 8 local rounds after switching back carried a
token produced only on the remote side; parent-side prompt grew from
10,888 to 13,127 bytes & Production evidence is available to local
repair without an explicit transfer step \\
Conversation locality & Parent read of the session returned 200; the
child's read of the same session ID returned 404 & History is not
migrated by binding; the parent remains the only holder \\
State non-migration & Remote writes existed only in the child workspace
(parent read: file not found); the bound round left one terminal
registered on the child and none on the parent & Artifacts, terminals,
and journals stay where they were created \\
Undo direction & Local undo reverted the post-switch local edit, then an
earlier local write, walking parent turns \texttt{261124\_231841} and
\texttt{261124\_231844}; child turns \texttt{261124\_231842} and
\texttt{261124\_231846} were untouched & The remote round is absent from
the local stack, so repeated undo walks past it and erases older local
work; production writes are not undoable from the development side \\
Workspace pin unresolvable at the bound site & A pin naming a
parent-only path while bound did not force that path and reported
success; the file landed at the executing environment's own workspace
root, verified out-of-band & The pin does not validate reachability, so
a targeted write can silently land elsewhere \\
Tool present only on the parent during binding & Search for a
parent-only string returned zero matches with a fallback engine rather
than an error & Catalog asymmetry surfaces as ``succeeded but empty'',
not as a diagnostic \\
Environment loss & Registering an unreachable environment was rejected
(502 in 2.1 ms); a bound child killed mid-session produced a tool-level
connection-refused error after 520.5 ms with no parent-side write, while
the registry still reported the environment online with no recorded
error & Failure is loud at execution and absent from status \\
Switching cost & Tool-free rounds 357.9 to 368.3 ms (median); rounds
with one tool call, mean 493.3 ms local, 509.0 ms bound same-host, 519.0
ms bound cross-host; the first two bound rounds were 2,105.2 and 2,178.9
ms & Warm switching costs tens of milliseconds; the cold cost is the
first bind \\
\end{longtable}
\endgroup

Three results change how the workflow should be operated. First,
recovery is distributed: each site keeps its own journal chain, so a
repair attempted after switching back cannot be rolled back by repeating
undo, and writes made during diagnosis are invisible to local undo. A
diagnosis round that mutates the production side must therefore be
recorded explicitly rather than trusted to the journal. Second, the two
degradations found are silent rather than loud: an unresolvable
workspace pin reports success, and a tool that exists only on the
unbound side returns an empty result instead of an error. Both were
detectable only by checking the other side out-of-band, which is the
same observability gap already visible in the online indicator. Third,
the cost is asymmetric in a way that suggests an optimization:
steady-state switching adds 10 to 26 ms per round at the tool-call
level, while the first bind of an environment cost about two seconds,
and across three runs 5 of 12 bound rounds exceeded 1.9 s. A warmed or
retained binding (cached remote tool lists and skill bodies) is the
indicated fix; we have not measured that fix itself.

The same distinction separates this workflow from the debugger below.
The debugger relocates the \emph{evidence source} inside one site, a
process on a host that already reached the parent; site switching
relocates \emph{every} execution operation, including file writes,
search, and journals. It also differs from a forwarded persistent
terminal, which relocates only the shell while relative paths continue
to resolve against the forwarded workspace pin. Neither comparison is a
measurement of which produces better diagnoses: with a scripted endpoint
the loop's mechanism is established, while the benefit of carrying
production evidence into local repair remains unmeasured.

\hypertarget{process-scale-execution-online_debugger}{%
\section{Process-Scale Execution:
online\_debugger}\label{process-scale-execution-online_debugger}}

\hypertarget{refining-the-execution-site}{%
\subsection{Refining the execution
site}\label{refining-the-execution-site}}

The debugger applies the same architectural split to an incident
process: reasoning belongs to the service, while frame access and
expression evaluation belong to the process that owns those objects. It
uses client-side handoff, not the infrastructure tunnel protocol. It
neither requires a child agentsvc instance inside the program nor opens
an additional inbound inspector port.

A Python host imports the accessory, starts it with project
configuration, and calls \texttt{report()} from an exception handler:

\begin{verbatim}
import online_debugger as debugger
debugger.start(project="example-service")
try:
    run_step()
except Exception as exc:
    debugger.report("run_step", exc)
\end{verbatim}

Reporting does not resume the failed call or alter the host's
exception-handling policy. Asynchronous dispatch allows the host to
continue its own control flow while diagnosis runs. A synchronous option
waits for diagnosis, which is useful for scripts but changes foreground
latency. A JavaScript accessory implements a corresponding handoff
workflow; the frame semantics described below are specifically Python
semantics.

\hypertarget{captured-evidence-and-inspection-tools}{%
\subsection{Captured evidence and inspection
tools}\label{captured-evidence-and-inspection-tools}}

The Python accessory walks the exception traceback and records the
exception chain, source locations, and truncated representations of
local variables. A traceback retains frame references after stack
unwinding, which makes later inspection possible. Locations come from
the traceback, while variable representations are generated when
\texttt{report()} captures them. This is near-exception evidence, not an
atomic snapshot: shared objects, globals, and still-active outer frames
may change.

Capture is limited to 12 frames, 30 locals per frame, and 200 characters
per variable representation. An LRU cache retains up to 512 frame
objects through strong references. The event table retains up to 32
entries and the dispatch queue holds up to 64; a full queue drops work.
These entry limits do not bound retained memory in bytes, because a
frame may retain a large object graph. Calling arbitrary \texttt{repr}
methods can also block or have side effects despite exception handling
and name-based exclusions.

The default fingerprint hashes exception type and frame function, file
basename, and line number. Including line numbers distinguishes
different failure sites along an otherwise identical call chain.
Repeated fingerprints increment counters without dispatching another
diagnosis in that process. Synchronous mode adds a persistent cooldown
record, normally 1,800 seconds, to reduce repeated diagnosis by
short-lived processes. Fingerprints are heuristic grouping keys:
collisions, source-line shifts, and suppressed variants remain possible.

\begingroup
\small
\setlength{\tabcolsep}{4pt}
\begin{longtable}[]{@{}
  >{\raggedright\arraybackslash}p{(\columnwidth - 4\tabcolsep) * \real{0.22}}
  >{\raggedright\arraybackslash}p{(\columnwidth - 4\tabcolsep) * \real{0.37}}
  >{\raggedright\arraybackslash}p{(\columnwidth - 4\tabcolsep) * \real{0.41}}@{}}
\caption{The debugger's six session-scoped client-side tools.
Capture-time evidence and query-time state answer different
questions.}\tabularnewline
\toprule\noalign{}
\begin{minipage}[b]{\linewidth}\raggedright
Tool
\end{minipage} & \begin{minipage}[b]{\linewidth}\raggedright
Evidence or operation
\end{minipage} & \begin{minipage}[b]{\linewidth}\raggedright
Temporal and authority boundary
\end{minipage} \\
\midrule\noalign{}
\endfirsthead
\toprule\noalign{}
\begin{minipage}[b]{\linewidth}\raggedright
Tool
\end{minipage} & \begin{minipage}[b]{\linewidth}\raggedright
Evidence or operation
\end{minipage} & \begin{minipage}[b]{\linewidth}\raggedright
Temporal and authority boundary
\end{minipage} \\
\midrule\noalign{}
\endhead
\bottomrule\noalign{}
\endlastfoot
\texttt{debugger\_status} & Process identity, uptime, event and dispatch
statistics & Current accessory state \\
\texttt{debugger\_events} & Retained events, counts, fingerprints,
diagnostic session IDs & Bounded event history \\
\texttt{debugger\_event} & One exception chain and captured frame
summaries & Capture-time evidence \\
\texttt{debugger\_frame} & Frame locals and source information by frame
ID & Captured evidence; retained objects are not frozen \\
\texttt{debugger\_threads} & Current stacks of process threads &
Query-time state \\
\texttt{debugger\_eval} & Evaluate an expression in a retained frame or
accessory scope & Live process execution; enabled by default; not a
sandbox \\
\end{longtable}
\endgroup

These tools are supplied as \texttt{client\_tools} on diagnostic
requests and are not added to the service's global registry. The service
advertises them to the model but cannot execute them. Frame-scoped
evaluation occurs in the incident process's diagnostic execution path;
it is not a mechanism that resumes execution on the original failing
thread.

\hypertarget{diagnostic-continuation}{%
\subsection{Diagnostic continuation}\label{diagnostic-continuation}}

\begin{figure}[htbp]
\hypertarget{fig:debugger}{%
\centering
\includegraphics[width=\linewidth,keepaspectratio]{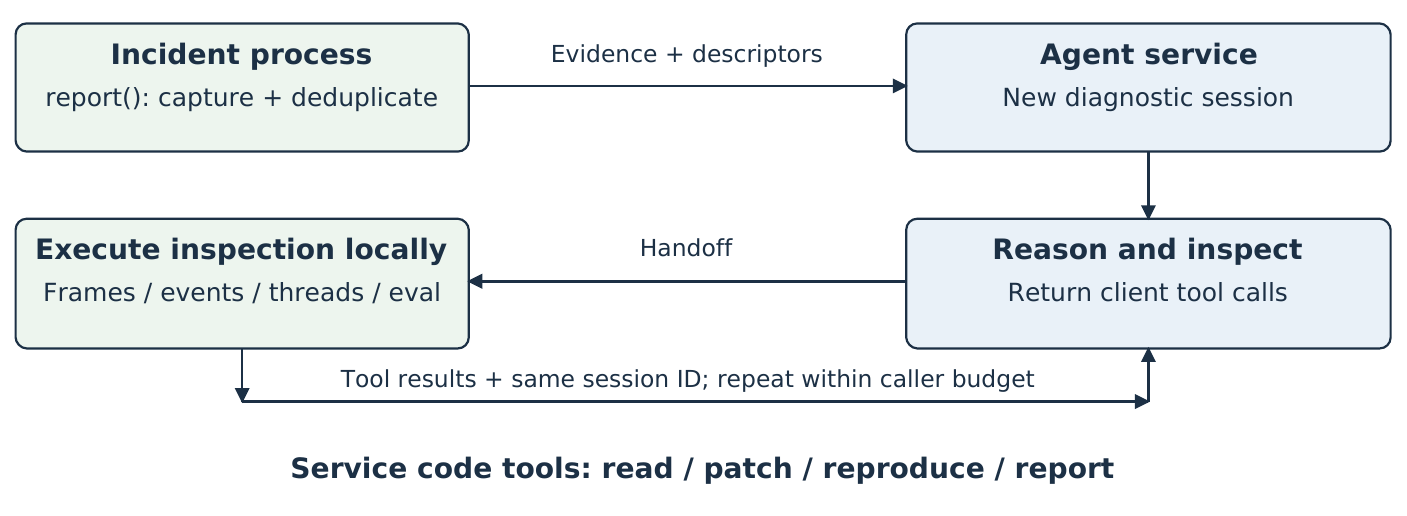}
\caption{Diagnostic continuation. Process-side inspection and
service-side code work occur at different sites. L4 requests
source-level reproduction evidence; L5 runtime activation and
failed-call recovery have no implementation.}\label{fig:debugger}
}
\end{figure}

The accessory creates a diagnostic session through \texttt{/v1/infer} or
its streaming variant, selecting a configured coding agent or model. The
request carries exception evidence, a workspace, and client-tool
descriptions. When inspection is needed, the service returns calls; the
accessory executes them and submits tool results with the same session
identifier (Figure \ref{fig:debugger}). Server-side code-reading and
editing tools can cooperate with process-side inspection in that
session.

The driver owns the continuation budget. The default is 80 handoffs. The
server's \texttt{max\_tool\_rounds}, normally 200, is per inference
request; a pure client-side handoff ends before consuming that counter,
and the next request starts a new budget. It therefore cannot bound an
entire handoff-driven diagnostic session. An HTTP request timeout of
1,200 seconds and a synchronous wait limit of 900 seconds are additional
limits, not one global termination guarantee. A daemon thread and a
non-throwing \texttt{report()} interface reduce some integration
failures but cannot isolate all resource consumption.

Streaming introduces a separate correctness obligation. Pending client
calls are the union of explicitly announced handoffs and declared calls
without tool receipts, deduplicated by call ID. A version containing
arguments is retained over an incomplete duplicate; calls already
answered are excluded. Otherwise, a partially announced parallel batch
can silently lose an inspection result or execute one again. The
accessory retries a transport-failed inference request once.
Deduplication within a returned batch and a retry do not establish
exactly-once processing across ambiguous network failure.

Workspace discovery starts at the innermost application frame, excluding
standard-library and installed-package locations, and otherwise uses the
current directory. An explicit workspace overrides discovery. A path
inferred on the incident machine may not exist at the service.
\texttt{DEBUGGER\_WORKSPACE} addresses source visibility only; it does
not move execution of verification commands. Shared storage, explicit
remote binding through an appropriate driver, or additional
source-access tools are needed for distributed deployments. The stock
diagnostic payload does not automatically select a \texttt{remote\_env}.

\hypertarget{source-repair-reproduction-and-recovery}{%
\subsection{Source repair, reproduction, and
recovery}\label{source-repair-reproduction-and-recovery}}

The diagnostic prompt asks the model to inspect evidence before forming
a root-cause hypothesis. It permits minimal source edits for a
sufficiently understood code defect, recommends retry or backoff for
transient external failures, and asks for advice rather than execution
for database writes or external configuration changes. The final report
has five sections: problem, root cause, changes, verification/risks, and
follow-up actions. These instructions are prompt-level policies, not
enforceable tool admission.

The default \texttt{verify\_fix=True} contract asks the agent to find or
construct a minimal reproduction that fails before the patch and passes
afterward. The failing reproduction should match the exception type and
top stack location. Commands, exit codes, and relevant outputs belong in
the report. If reproduction is unavailable, the report must explicitly
state that the change is unverified and why. \texttt{DEBUGGER\_VERIFY=0}
removes the action requirement but still requires disclosure that
verification was not performed.

This L4 task uses existing code-editing and shell tools; no separate
verification component is required to request it. The implementation
test checks that the prompt contains the evidence requirement. It does
\textbf{not} show that the model follows it, that a generated
reproduction is adequate, or that a reported passing test establishes
correctness. Independent hidden tests and recorded tool receipts are
needed to assess those outcomes.

\begin{figure}[htbp]
\hypertarget{fig:stages}{%
\centering
\includegraphics[width=\linewidth,keepaspectratio]{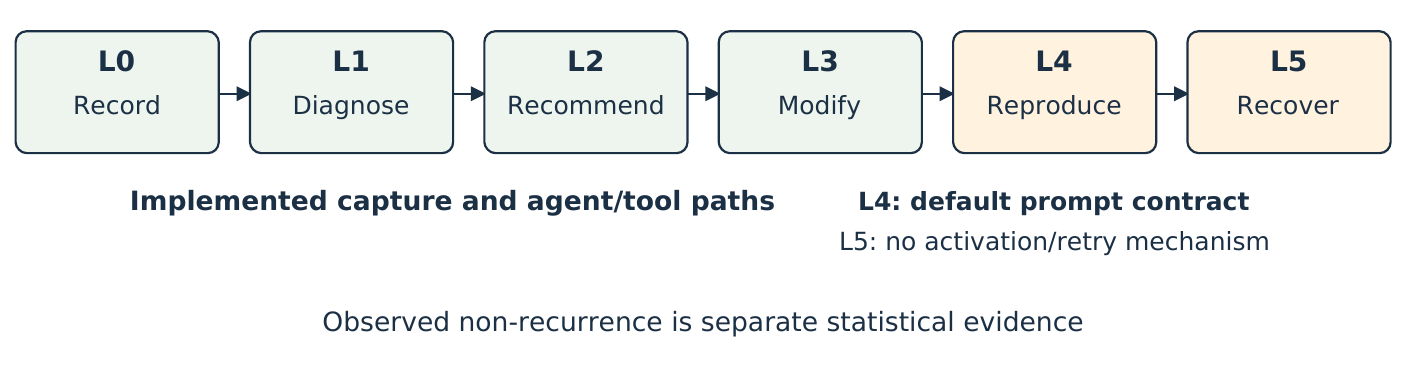}
\caption{Repair stages and their evidence boundaries. L4 is a default
prompt requirement whose execution and adequacy must be assessed. Source
changes and passing reproductions do not activate patches or restore
failed calls.}\label{fig:stages}
}
\end{figure}

Figure \ref{fig:stages} and Table 10 distinguish the stages. L0--L3
provide capture, diagnosis, recommendations, and source modification
paths. L4 is enabled as a default behavioral contract. L5 requires
mechanisms for activating a patch in the running program and recovering
the failed call; these are absent. The taxonomy is local to this paper
and is not an external maturity standard.

\begingroup
\small
\setlength{\tabcolsep}{4pt}
\begin{longtable}[]{@{}
  >{\raggedright\arraybackslash}p{(\columnwidth - 4\tabcolsep) * \real{0.2}}
  >{\raggedright\arraybackslash}p{(\columnwidth - 4\tabcolsep) * \real{0.4}}
  >{\raggedright\arraybackslash}p{(\columnwidth - 4\tabcolsep) * \real{0.4}}@{}}
\caption{Repair stages, implementation status, and permissible
claims.}\tabularnewline
\toprule\noalign{}
\begin{minipage}[b]{\linewidth}\raggedright
Stage
\end{minipage} & \begin{minipage}[b]{\linewidth}\raggedright
Mechanism in this implementation
\end{minipage} & \begin{minipage}[b]{\linewidth}\raggedright
Supported claim
\end{minipage} \\
\midrule\noalign{}
\endfirsthead
\toprule\noalign{}
\begin{minipage}[b]{\linewidth}\raggedright
Stage
\end{minipage} & \begin{minipage}[b]{\linewidth}\raggedright
Mechanism in this implementation
\end{minipage} & \begin{minipage}[b]{\linewidth}\raggedright
Supported claim
\end{minipage} \\
\midrule\noalign{}
\endhead
\bottomrule\noalign{}
\endlastfoot
L0 Record & Exception capture, event and frame retention & Bounded
incident evidence is available. \\
L1 Diagnose & Agent session with inspection tools & The agent can form
evidence-informed hypotheses. Accuracy is unmeasured. \\
L2 Recommend & Disposition instructions and report & The agent can
propose action or abstention. Compliance is unmeasured. \\
L3 Modify & Code tools with access to the intended source & Source files
can be patched; loaded code may remain unchanged. \\
L4 Reproduce & Default failing-then-passing evidence requirement &
Verification is requested using existing tools; completion and adequacy
are not enforced. \\
L5 Activate and recover & No general mechanism & Runtime patch
activation and failed-call recovery remain gaps. \\
\end{longtable}
\endgroup

Ordinary Python functions already loaded in memory do not change when
their source file is rewritten. Restart, controlled reload, or
host-specific update protocols may activate later code, each with its
own state and reference hazards. Replaying the failed call also requires
knowing whether it already produced side effects. The prompt forbids
independently restarting the host or using in-process reload to
manufacture apparent recovery. A successful source-level reproduction
cannot establish that the incident process has recovered.

Reproductions are also limited by captured information. Truncated locals
and excluded connection objects may omit exactly the external state
needed to reproduce a failure. Concurrent schedules, transient rate
limits, and mutable services are not reconstructed by a traceback. A
generated test can provide useful evidence without replaying the
incident exactly. Subsequent non-recurrence is observational evidence,
influenced by exposure and suppression policies; it is not a proof of
repair.

\hypertarget{integration-and-security-boundary}{%
\subsection{Integration and security
boundary}\label{integration-and-security-boundary}}

An earlier accessory exposed an HTTP inspector in the incident process
and required a diagnostic agent to query it through a shell and
\texttt{curl}. Client-side handoff replaces that endpoint with
structured calls over the program's outbound inference exchange. It
removes a listening port, port conflicts, and inspector-specific CLI
integration. It preserves a powerful authority channel: model-generated
requests still control operations in the incident process.

In particular, \texttt{debugger\_eval} can call functions with side
effects. \texttt{DEBUGGER\_EVAL=0} disables it; local execution is
neither read-only isolation nor a sandbox. Locals can contain secrets,
and truncation or a name blacklist does not provide adequate general
redaction. Retained frames, representation methods, diagnostic requests,
and model-generated tests can consume resources or perturb a program.

The verification prompt requests reproductions without real-data writes
or external calls, using disposable copies where appropriate. A
temporary directory alone does not enforce those restrictions. Running
generated tests with production credentials expands authority unless the
host supplies an enforceable boundary. This implementation provides
service authentication but lacks per-tool approval and sandbox policies.
Consequently, it offers a mechanism for bringing LLM-assisted diagnosis
and source repair into ordinary exception handling; autonomous
production healing remains an unvalidated goal.

\hypertarget{comparison-and-related-work}{%
\section{Comparison and Related
Work}\label{comparison-and-related-work}}

\hypertarget{scope-and-method-of-comparison}{%
\subsection{Scope and method of
comparison}\label{scope-and-method-of-comparison}}

We compare concrete mechanisms with Hermes Agent, Pi, Codex, and Claude
Code because they expose different choices in agent delivery,
extensibility, and execution \citep{hermes, pi, codex, claude}. The
documentation review used repository material collected on November 7,
2026. Coverage was uneven: Hermes and Pi were represented mainly by
README material, Codex by its README and documentation index rather than
every linked page, and Claude Code by a retained
system-prompt/tool-description snapshot dated May 9, 2026. That snapshot
is historical, not a public reproducible substitute for current official
documentation. Current Claude Code claims therefore remain provisional.

Table 11 compares architectural dimensions without interpreting missing
evidence as missing capability. An integrated counterpart is
\emph{unverified} when the reviewed material does not establish the full
contract. This leaves room for extensions, external integrations, and
features outside the reviewed versions. No same-task product benchmark
was performed.

\begingroup
\small
\setlength{\tabcolsep}{4pt}
\begin{longtable}[]{@{}
  >{\raggedright\arraybackslash}p{(\columnwidth - 10\tabcolsep) * \real{0.14}}
  >{\raggedright\arraybackslash}p{(\columnwidth - 10\tabcolsep) * \real{0.2}}
  >{\raggedright\arraybackslash}p{(\columnwidth - 10\tabcolsep) * \real{0.165}}
  >{\raggedright\arraybackslash}p{(\columnwidth - 10\tabcolsep) * \real{0.165}}
  >{\raggedright\arraybackslash}p{(\columnwidth - 10\tabcolsep) * \real{0.165}}
  >{\raggedright\arraybackslash}p{(\columnwidth - 10\tabcolsep) * \real{0.165}}@{}}
\caption{Item-by-item architectural comparison. ``Unverified'' describes
the reviewed evidence, not product absence. Product claims require
version-pinned official verification before submission.}\tabularnewline
\toprule\noalign{}
\begin{minipage}[b]{\linewidth}\raggedright
Dimension
\end{minipage} & \begin{minipage}[b]{\linewidth}\raggedright
agentsvc
\end{minipage} & \begin{minipage}[b]{\linewidth}\raggedright
Hermes Agent
\end{minipage} & \begin{minipage}[b]{\linewidth}\raggedright
Pi
\end{minipage} & \begin{minipage}[b]{\linewidth}\raggedright
Codex
\end{minipage} & \begin{minipage}[b]{\linewidth}\raggedright
Claude Code
\end{minipage} \\
\midrule\noalign{}
\endfirsthead
\toprule\noalign{}
\begin{minipage}[b]{\linewidth}\raggedright
Dimension
\end{minipage} & \begin{minipage}[b]{\linewidth}\raggedright
agentsvc
\end{minipage} & \begin{minipage}[b]{\linewidth}\raggedright
Hermes Agent
\end{minipage} & \begin{minipage}[b]{\linewidth}\raggedright
Pi
\end{minipage} & \begin{minipage}[b]{\linewidth}\raggedright
Codex
\end{minipage} & \begin{minipage}[b]{\linewidth}\raggedright
Claude Code
\end{minipage} \\
\midrule\noalign{}
\endhead
\bottomrule\noalign{}
\endlastfoot
Delivery and composition & HTTP service, SDK, browser and TUI; runtime
registries & Personal/server agent with messaging gateway and
configuration & Extensible coding harness, RPC and SDK & CLI, IDE and
hosted task forms & CLI/IDE and SDK; historical tool snapshot \\
Execution-site mechanism & One child per request; file, terminal, MCP
and journal integration & Documented local, container, SSH and sandbox
terminal backends & Extensible execution through harness integrations &
Local and hosted execution forms & Local tools and remote integration
forms \\
Managed private fleet and reverse transport & Roster, status snapshots,
push updates; direct and reverse paths & Full counterpart unverified &
Full counterpart unverified & Full counterpart unverified & Full
counterpart unverified \\
Caller-owned tool continuation & Session-scoped descriptors, caller
execution, same-session results & Equivalent integrated contract
unverified & RPC/extensions documented; equivalent contract unverified &
Equivalent integrated contract unverified & Interactive tools observed;
general contract unverified \\
Exception-triggered in-process inspection & Included accessory with six
process-side tools & Integrated counterpart unverified & Integrated
counterpart unverified & Integrated counterpart unverified & Integrated
counterpart unverified \\
Features not claimed as unique & Skills, subagents, models, persistence,
shell tools & Skills, learning, subagents and scheduling documented &
Skills and extensions; deliberately minimal built-in policy & Coding
tools, skills and sandbox/approval documentation & Agent, skill,
planning and permission mechanisms \\
\end{longtable}
\endgroup

The strongest difference candidates are the \emph{combination} of a
standard-library service core, centrally operated child environments,
consistent request and session semantics across transports, and
caller-executed process inspection. HTTP calls, reverse tunnels,
snapshots, client-side functions, and remote credentials are established
techniques. Their integration is the contribution investigated here; the
review does not establish priority or uniqueness.

Several dimensions favor alternative systems. Hermes documents learning
and skill-creation workflows that agentsvc does not automatically
provide. Pi deliberately offers a minimal extensible harness. Codex
documents sandbox and approval policies absent from agentsvc. Claude
Code's permission mechanisms illustrate a control layer beyond
service-level authentication. Such differences affect deployment
suitability and cannot be summarized by counting supported features.
Likewise, a standalone binary may be easy to install despite having a
different implementation dependency model.

\hypertarget{research-context}{%
\subsection{Research context}\label{research-context}}

ReAct studies the interleaving of reasoning and action, Reflexion
studies language-based feedback, and Toolformer studies learning to use
tools \citep{react, reflexion, toolformer}. This paper concentrates on
runtime ownership and placement rather than proposing a new model
policy. A runtime can use those policies while independently deciding
where actions execute.

SWE-agent and SWE-bench make repository-level software engineering
measurable through agent interfaces and testable issue instances
\citep{sweagent, swebench}. The debugger contributes a different input
source: runtime exception evidence and access to retained frames. It can
complement coding agents rather than replace them. Automated program
repair, including GenProg, supplies a longer history of patch generation
and test-based validation \citep{genprog}. Passing a reproduction
remains weaker than establishing semantic correctness, especially when
tests are produced by the same agent that writes the patch.

Autonomic computing motivates monitoring, analysis, planning, execution,
and knowledge in feedback systems \citep{autonomic}. The debugger
supplies bounded incident knowledge and uses an LLM for analysis and
planning, with source edits as one possible action. The separation
between requested verification and unimplemented runtime recovery
prevents equating that partial path with a complete self-healing loop.

MCP standardizes tool interaction, and Agent Skills standardizes
reusable skill descriptions \citep{mcp, skills}. agentsvc implements
these interfaces alongside execution placement. It does not claim that
remote tool use, shared tool descriptions, or progressive skill
disclosure originated here. General remote procedure calls, reverse
tunnels, and debugger integrations also predate this system; our bounded
product review is not an exhaustive prior-art search across distributed
runtimes and observability platforms.

\hypertarget{evidence-limitations-and-evaluation-agenda}{%
\section{Evidence, Limitations, and Evaluation
Agenda}\label{evidence-limitations-and-evaluation-agenda}}

\hypertarget{measurement-program-and-validity-protocol}{%
\subsection{Measurement program and validity
protocol}\label{measurement-program-and-validity-protocol}}

The quantitative claims in Sections 2 through 5 come from a program of
seven experiment families run against the same implementation baseline,
governed by one validity contract written before any of them (artifact
labels E1, E2, E3, E3B, E3C, E4, E6, E7, E9). The contract has 14 checks
in five groups: environment isolation, routing proof, provenance,
timestamp discipline, and the declared status of one candidate finding.
Two of its rules did real work. A run whose routing proof fails is void
rather than partial, because a value held only at the far end must be
retrievable only through the measured route and a local witness must be
unreadable there: a silent fallback then appears as a wrong value
instead of a silent pass. And failures are classified: a system failure
under a correct configuration is a result that stays in the denominator,
while a harness misconfiguration marks the run invalid, is rerun, and
keeps the bad record.

\begingroup
\small
\setlength{\tabcolsep}{4pt}
\begin{longtable}[]{@{}
  >{\raggedright\arraybackslash}p{(\columnwidth - 6\tabcolsep) * \real{0.16}}
  >{\raggedright\arraybackslash}p{(\columnwidth - 6\tabcolsep) * \real{0.26}}
  >{\raggedright\arraybackslash}p{(\columnwidth - 6\tabcolsep) * \real{0.28}}
  >{\raggedright\arraybackslash}p{(\columnwidth - 6\tabcolsep) * \real{0.3}}@{}}
\caption{The measurement program at a glance. Every quoted number
belongs to an archived run and is covered by a forward assertion in that
experiment's checker; none of these families was model
driven.}\tabularnewline
\toprule\noalign{}
\begin{minipage}[b]{\linewidth}\raggedright
Family
\end{minipage} & \begin{minipage}[b]{\linewidth}\raggedright
Question
\end{minipage} & \begin{minipage}[b]{\linewidth}\raggedright
Scale as run
\end{minipage} & \begin{minipage}[b]{\linewidth}\raggedright
Strongest measured result
\end{minipage} \\
\midrule\noalign{}
\endfirsthead
\toprule\noalign{}
\begin{minipage}[b]{\linewidth}\raggedright
Family
\end{minipage} & \begin{minipage}[b]{\linewidth}\raggedright
Question
\end{minipage} & \begin{minipage}[b]{\linewidth}\raggedright
Scale as run
\end{minipage} & \begin{minipage}[b]{\linewidth}\raggedright
Strongest measured result
\end{minipage} \\
\midrule\noalign{}
\endhead
\bottomrule\noalign{}
\endlastfoot
Routing and transport (E1) & What does routing one tool call to a child
cost & 3 routing runs (63 samples per cell) and 4 tunnel runs (12 per
cell per run) & 1.8 to 2.2 round-trips per call; forwarding +0.84 ms on
a 4 KiB read \\
Persistent terminals (E3, E3B, E3C) & What a persistent terminal buys,
and where it costs & 12 tasks by 4 arms (172 trials); 128 cross-path
trials; 196 timed calls on a deployed instance & Capability, not
latency; completion strategy 4.2 to 19.6x; 54 flat steps after the
fix \\
Interactive layers (E4) & Cost and correctness when crossing a layer & 4
arms by 7 chains, 56 trials and 404 call rows over 2 runs & Entry tax
17.6x and 383x; wrong-host writes in 8 of 8 trials on one chain \\
Routing-arm semantics (E2) & Is a tool the same tool after re-routing &
6 arms by 14 tasks by 2 batches: 168 task-runs, 144 checks & Context,
not routing, carries semantics \\
Injected faults (E6) & What failures mean and what recovery costs & 9
classes by 10 tasks, 264 rows, plus 3 stages & One endpoint, two
executor paths, invisible to the caller \\
Authorization and context (E7) & Is forwarded context trustworthy & 124
rows, 11 checks, 285 archived checksums & Context is not
authentication \\
Execution-site switching (E9) & What follows a site switch inside one
session & 16 checks, 46 observation rows, 4 timing repetitions &
Switching is reversible and cheap; diagnosis carries back \\
\end{longtable}
\endgroup

Reproduction was checked rather than assumed. The interactive-ladder
study ran 28 trials twice with no disagreement in cell pass rates and a
maximum single-step median deviation of 7.4\%; the routing study merged
three and four runs only after each had passed its routing proof; the
authorization study ran the same task family twice after an earlier
variant was invalidated; and the switching pilot keeps three superseded
runs alongside its reference run, each with the reason for supersession
recorded. Artifact integrity is checked by hash, with path conventions
documented per archive because the wrong working directory produces a
failure that looks like data corruption.

Two negative controls are reported because they changed what the numbers
mean. In the interactive study's first run the driver failed to forward
two keyword arguments, so the prompt pattern and the timeout were never
sent and the four completion arms were in fact one arm: all 39
arm-to-arm ratios landed between 0.99 and 1.04. The run was marked
invalid and rerun, and it survives as a demonstration that arm
comparisons silently degenerate when a driver bug is invisible. In the
switching pilot, an edit call missing a required argument produced a
\texttt{TypeError} that the harness first read as a product failure; the
affected check was invalidated and rerun, which is the second half of
the failure-classification rule.

The program also produced three methodological results worth stating
because they are not specific to this system. Silent degradation, not
error, was the dominant failure form: state loss, the executor-path
fork, an unresolvable forwarded pin, a parent-only tool called under a
remote binding, and an unrecoverable stateless journal handle all
returned success or an empty result. The root cause was usually an
observability gap: no mode field for the path fork, no exposed admission
count, creation failures logged only at debug level, and journal turn
keys visible only in the executing side's files. And the measurement
apparatus is subject to the same class of problem: a harness's own
residue count was wrong because the counter could only rise, and three
hidden correctness criteria in the repair-fixture set had been quietly
loosened, which is why those criteria are now hash-locked and negatively
tested (a plausible but wrong patch must pass the visible tests and fail
the hidden ones).

Two gaps in the program are stated plainly because they bound what the
paper can claim. The repair-ablation family has fixtures and locked
criteria only: 10 incidents across 6 root-cause classes with 119
checksummed entries, a model endpoint measured for connectivity, and
zero model-driven rows. The version-matrix, clean-machine deployment,
and context-compression family has no data at all.

\hypertarget{implementation-and-regression-evidence}{%
\subsection{Implementation and regression
evidence}\label{implementation-and-regression-evidence}}

The baseline verification record describes a runtime of 42 Python files
and approximately 44,000 lines, a browser frontend of approximately
37,000 lines, and 98 Python test files at the fixed baseline. These
counts establish implementation scope only. They are not evidence of
minimal code size, coverage, quality, or comparative maturity. The
AST-based dependency test checks the core's optional import allow-list;
it does not audit the dependencies of external tools.

Table 13 reports the baseline verification record. The targeted
selections overlap with the full suite and must not be added to its
count. The reported environment was Linux, Python 3.13.11, pytest 9.0.3,
with \texttt{HYPOTHESIS\_PROFILE=fast}. The JavaScript self-check used
Node v24.13.0 and scripted transport without a model service. These are
recorded regression checks; the controlled measurements reported in
Sections 3, 4, and 7 are separate from them, and the model-driven repair
study remains unrun.

\begingroup
\small
\setlength{\tabcolsep}{4pt}
\begin{longtable}[]{@{}
  >{\raggedright\arraybackslash}p{(\columnwidth - 4\tabcolsep) * \real{0.36}}
  >{\raggedleft\arraybackslash}p{(\columnwidth - 4\tabcolsep) * \real{0.15}}
  >{\raggedright\arraybackslash}p{(\columnwidth - 4\tabcolsep) * \real{0.49}}@{}}
\caption{Regression evidence recorded for the fixed implementation
baseline. No run measures model repair accuracy or deployment-scale
performance.}\tabularnewline
\toprule\noalign{}
\begin{minipage}[b]{\linewidth}\raggedright
Recorded run
\end{minipage} & \begin{minipage}[b]{\linewidth}\raggedleft
Result
\end{minipage} & \begin{minipage}[b]{\linewidth}\raggedright
What it checks
\end{minipage} \\
\midrule\noalign{}
\endfirsthead
\toprule\noalign{}
\begin{minipage}[b]{\linewidth}\raggedright
Recorded run
\end{minipage} & \begin{minipage}[b]{\linewidth}\raggedleft
Result
\end{minipage} & \begin{minipage}[b]{\linewidth}\raggedright
What it checks
\end{minipage} \\
\midrule\noalign{}
\endhead
\bottomrule\noalign{}
\endlastfoot
Tunnel suite & 56 passed & Tunnel mechanisms, including replay target,
TLS and error responses \\
Debugger and client-tool suites & 40 passed & Handoff completeness,
budgets, identity, workspace and prompt contract \\
Remote execution and remote orchestration suites & 61 passed & Wrapping,
context and parent-side orchestration behavior \\
Dependency and HTTPS suites & 70 passed & Optional-import constraint,
certificate and authentication paths \\
Full Python regression & 1,969 passed & Repository regression checks;
targeted runs above are subsets \\
Offline JavaScript self-check & 31 passed & Scripted transport and
debugger contract behavior \\
\end{longtable}
\endgroup

The most informative tests protect cross-boundary obligations: resolving
the child's actual listening target, using TLS when appropriate,
returning local exceptions as responses, preserving all pending client
calls, and separating server-round from handoff budgets. A prompt
assertion protects the presence of a verification instruction. It does
not validate agent behavior. Wall-clock-sensitive tests can also vary
under load; a successful regression run is not evidence of timing
stability or throughput.

\hypertarget{missing-measurements-and-threats-to-validity}{%
\subsection{Missing measurements and threats to
validity}\label{missing-measurements-and-threats-to-validity}}

The measurement program described above bounds mechanism-level cost and
semantics for routing, completion decisions, re-routing, injected
faults, authorization, and site switching; it does not measure benefit.
Remote overhead under load remains open beyond 16 concurrent calls, and
reconnect distributions, throughput under mixed load, round-trip-time
sweeps across network conditions, and behavior on heterogeneous networks
are unmeasured; the routing figures come from one parent-child pair on
one network, and the cross-host switching figure from one deployed
endpoint. Repair accuracy, mis-edit rates, and reduction in
incident-resolution time remain unmeasured, and the switching loop's
benefit is likewise unmeasured: because that run used a deterministic
endpoint, it establishes that production evidence reaches the local
repair round, not that using it improves repair outcomes. No arm in the
program was model driven, so nothing here shows that a model exploits
the capabilities that pass the mechanical tests. Default frame sizes and
worker counts are choices awaiting measurement. We also have not shown
that centralized management reduces total operational cost, that the
equivalence result obtained for one tool family on two transports
extends to every local capability or to other deployments, or that
context compression preserves the information needed for long tasks.

Internal validity is limited by the use of unit and regression checks,
often with controlled transports, rather than production workloads, and
by the measurement program's pilot scale: fast-cell medians are
repetition-stable (a maximum 7.4\% single-step median deviation over two
runs of 28 trials, 10 ms of cell spread across three routing runs),
while slow cells and tails are not, and every figure was produced by
scripted drivers on a small set of Linux hosts of two distributions with
no model choosing strategies. Construct validity requires separating
available mechanisms from achieved outcomes: a tool receipt, a plausible
explanation, a written patch, a passing reproduction, and a recovered
process are distinct observations. External validity is constrained by
one project's implementation, a Linux verification environment,
restricted legacy-Python roles, and uneven competitor documentation. The
author's design discussions explain intent but provide no independent
evaluation.

Engineering limits include a threaded HTTP server without demonstrated
load scaling, tool-definition drift without general negotiation, an
online indicator that omits local replay health, no tool-level admission
or multi-tenant isolation, and retained-frame resource risks. The
measurements above add three specifics to that list: a registered
environment can be reported online after its process has gone, the
choice between the two executor paths is visible only at debug level,
and a forwarded workspace pin that no longer resolves falls back to the
executing side's default workspace with only a warning line. Service
authentication and journals address only parts of these problems.
Default local TLS verification and generated-code execution require
deployment-specific trust decisions. The absence of L5 makes runtime
recovery claims premature even if an L4 reproduction passes.

\hypertarget{three-priority-studies}{%
\subsection{Three priority studies}\label{three-priority-studies}}

\textbf{Remote transport overhead (partly completed at the tool-call
level).} Compare local execution, direct HTTP, reverse WebSocket, and a
CONNECT-proxied reverse path, with an optional well-configured SSH
baseline for comparable operations. Use deterministic tools first to
separate runtime overhead from model latency, then complete tasks. Vary
network RTT, payload size around 256 KiB framing and 8 MiB spilling, and
concurrency at 1, 8, and 16. Report p50/p95/p99 latency, throughput, CPU
and memory, reconnection time, failure propagation, and stranded child
work after parent timeout. Do not infer performance from the configured
limits. \emph{Run:} the local, direct, proxied, and reused-tunnel arms
at median and 95th percentile for a 4 KiB to 256 KiB ladder, with
concurrency at 1, 8, and 16 for one operation (Section 4).
\emph{Outstanding:} p99 and tail behavior under load, reconnection time,
CPU and memory, full round-trip-time sweeps, and the SSH baseline.

\textbf{Capability alignment (completed as a pilot for six of these
surfaces).} Run identical file-edit, journal, terminal, skill, MCP, and
delegated-agent tasks locally and through both remote paths. Fix tool
configuration, model, prompts, and budgets. Assert effects in the
intended filesystem and absence of unintended parent-local changes, not
merely equal model answers. Ablate the context header and test missing
workspaces, unavailable children, schema drift, and control-plane-only
health. This study would establish the conditions under which the shared
interfaces preserve task semantics. \emph{Run:} file edit, journal,
terminal, workspace and search visibility, skill fetch, catalogue
comparison, and delegation across six routing arms, asserting effects in
the intended filesystem rather than equal model answers (168 task-runs
and 144 checks, Section 3). \emph{Outstanding:} MCP tools,
delegated-agent task behavior, schema drift, the context-header
ablation, and any replication outside the two transports and two hosts
used here.

\textbf{Incident-evidence ablation (not started; fixtures ready).} Hold
the model, source snapshot, and code tools fixed. Compare A: traceback
only; B: traceback plus captured locals; C: B plus event/frame/thread
queries; D: C plus expression evaluation. Include deterministic defects,
state-dependent incidents, and transient external errors for which
abstaining from a patch is appropriate. Measure root-cause accuracy,
independent hidden-test pass rate, regressions and mis-edits, time,
inference cost, and foreground overhead. Record proposed, written,
reproduction-tested, independently validated, and activated patches
separately. Without host-supported activation, measure source repair
only. Repeat runs, publish failures and uncertainty intervals, and keep
the repair model from serving as the sole correctness judge.
\emph{Ready:} 10 incident fixtures across 6 root-cause classes with
visible tests and hash-locked hidden criteria, negatively validated so
that a plausible wrong patch passes the visible tests and fails the
hidden ones, plus a connectivity probe of the intended model endpoint.
\emph{Outstanding:} every model-driven arm, and therefore every
repair-quality number.

The outstanding parts of these studies should pin artifact and
competitor versions and retain raw traces. Enforceable tool policies,
validation at the correct site, data-plane health probes, and
host-specific activation protocols are engineering priorities alongside
measurement. The remaining items are future work; the pilot results
above settle mechanism behavior, not these questions.

\hypertarget{conclusion}{%
\section{Conclusion}\label{conclusion}}

Linglong Agent Service makes execution site an explicit part of agent
composition. A parent retains reasoning orchestration and session
management while compatible execution tools act in a selected child
environment. Portable context and common application handlers connect
direct HTTP, reverse tunnels, terminals, workspaces, and operational
updates without requiring a second model-facing tool vocabulary.

online\_debugger refines the same placement principle to a program that
owns transient exception evidence. Client-side handoff allows an
external agent to inspect that evidence and work on source code without
an additional inbound inspector. Its default reproduction requirement
distinguishes source-level evidence from unimplemented runtime
activation and failed-call recovery.

The contribution this version supports is an implemented architecture
with inspectable contracts, a regression record at a fixed baseline, and
a mechanism-level measurement program whose validity contract is
reusable independently of this system. Three of our own design
assumptions failed inside those measurements rather than in review: that
how a tool decides a step is finished is an implementation detail, worth
a factor of 4.2 to 19.6; that a workspace pin defines where a write
lands, since an unresolvable pin reported success while the file landed
in the executing environment's own workspace; and that a degraded
session stays visibly degraded, where ten live sessions were followed by
answers in 2.8 to 4.1 ms carrying no marker in the response and no trace
in the service log. Position is the recurring blind spot: all eight
trials of one nested chain wrote their artifact on the wrong host while
every in-band signal reported success, and the entering call behind an
8.2 s link establishment cost 383x what the steady step on that same
link cost. What the measurements establish is cost and semantics, not
benefit: deployment value, capability alignment, and diagnostic
effectiveness remain open to the controlled studies described above. The
source is available under the MIT license at the fixed repository
baseline \citep{agentsvc}.

\bibliographystyle{plainnat}
\bibliography{references}
\end{document}